%% file: main.tex
\documentclass{article}

\PassOptionsToPackage{numbers,square,sort&compress}{natbib}
\usepackage[final,main]{neurips_2026}
\setcitestyle{numbers,square,comma}
\usepackage[utf8]{inputenc}
\usepackage[T1]{fontenc}
\usepackage{amsmath}
\usepackage{amssymb}
\usepackage{amsthm}

\newtheorem{corollary}{Corollary}
\newtheorem{theorem}{Theorem}
\usepackage{booktabs}
\usepackage{array}
\usepackage{microtype}
\usepackage{graphicx}
\usepackage[utf8]{inputenc}
\usepackage{hyperref}
\hypersetup{hypertexnames=false}
\usepackage{url}
\usepackage{nicefrac}
\usepackage{xcolor}
\usepackage{multirow}
\usepackage[capitalize,noabbrev]{cleveref}
\usepackage{tikz}
\usepackage{pgfplots}
\usepackage{float}

\title{Physics-Informed Functional Tucker Method with RKHS Factors for Sparse Spatiotemporal Reconstruction}

\author{
Zhengyang Liu$^{1}$ \and
Jianguo Huang$^{2}$ \and
Junxi Hu$^{1}$ \and
Yucheng Mi$^{1}$ \and
Haojie Cai$^{1}$ \and
Wenxuan Shen$^{3}$ \and
Zhikun Zhang$^{1}$ \and
Renfeng Peng$^{5,*}$ \and
Guangtao Zhang$^{1,4,*}$ \and
Zhiqiang Liu$^{4,}$\thanks{Corresponding author.} \\
$^1$SandGold AI Research, Guangzhou, China \qquad \\
$^2$Nanyang Technological University, Singapore \qquad \\
$^3$Westlake University, Hangzhou, China \qquad \\
$^4$Southern University of Science and Technology, Shenzhen, China \qquad\\
$^5$Chinese Academy of Sciences, Beijing, China \qquad\\
{\tt\small {pengrenfeng@lsec.cc.ac.cn}; {tao@sgd-ai.com}; {liuzq@sustech.edu.cn}}
}

\usepackage{xcolor}

\begin{document}
\maketitle
\raggedbottom

\begin{abstract}

\begingroup
Reconstructing physical dynamics from fixed sparse sensors and missing frames is a fundamental challenge in scientific research. Sparsity leaves large regions of the field unconstrained, making the inverse problem ill-posed: a model may fit the observed sensors while producing inaccurate off-sensor structures or unreliable predictions between support frames. To this end, we propose PhysFTM---a physics-informed functional Tucker model that parameterizes the Tucker mode factors in reproducing kernel Hilbert spaces (RKHSs) and incorporates PDE constraints through finite-difference residuals evaluated at off-sensor collocation points. These residuals provide additional physical information in regions without direct observational supervision. PhysFTM first reconstructs continuous spatial fields at support frames via an alternating minimization scheme, optimizing the Tucker core tensor and the RKHS-based factors separately. To enable missing-frame reconstruction, PhysFTM constructs a continuous trajectory in Tucker-core space, which is decoded by the learned spatial RKHS-FTM representation to recover the full spatiotemporal field. Experiments on Allen--Cahn, Navier--Stokes, three-dimensional Heat equations, and a three-dimensional gridded Copernicus ocean reanalysis demonstrate accurate 2D/3D reconstruction under two missing-frame protocols. Code is available at https://github.com/liusirzzz/PhysFTM.
\endgroup

\end{abstract}

\input{sections/01_introduction}

\input{sections/02_related_work}

\input{sections/03_methodology}

\input{sections/04_experiments_neurips}
\input{sections/05_conclusion}

\section*{Acknowledgement}
This work was supported by the National Natural Science Foundation of China (Grant No. 42450181) and by a grant from the Research Grants Council of the Hong Kong Special Administrative Region, China (Grant No. AoE/P-601/23-N).

\bibliographystyle{plainnat}
\bibliography{references}

\clearpage

\appendix
\input{sections/06_appendix}

\clearpage

\end{document}

%% file: sections/01_introduction.tex
\section{Introduction}

\begingroup
Reconstructing spatiotemporal fields aims to recover full physical fields from a small set of fixed sensor locations, which is a fundamental problem in scientific research~\citep{manohar2018sparse,erichson2020shallow,xu2023practical,xu2023spatiotemporal,gao2024riemannian,zhu2025diffuse,zhang2026data}. Such sparsity leaves most locations and time instants unconstrained, making the inverse problem severely underdetermined: fine-scale structures such as vortices, fronts, and submesoscale processes may remain unresolved, and a model may fit the observed sensors while producing nonphysical off-sensor fields or unreliable missing-frame predictions~\citep{xu2023pinns}. This motivates reconstruction methods that are accurate at sensor locations while improving physical consistency on the missing-frame off-sensor locations.
\endgroup

\begingroup
Data-driven reconstruction methods have made substantial progress in sparse physical-field recovery~\cite{LiSABNN,zhanga2025parameter,zhang2026solving}, including sensor-to-field decoders, neural implicit representations, low-rank functional tensors, and generative latent models~\cite{chen2025sdift,han2025climate,hana2026generative,han2026benchmarkingphysicsinformedtimeseriesmodels}. A common paradigm is to first reconstruct fields or latent states from sparse measurements and then use interpolation, recurrent models, or generative priors to infer the temporal evolution~\cite{chen2025sdift}. Functional Tucker models offer compact representations of continuous fields by combining mode-wise factor functions with a low-dimensional core tensor~\citep{tucker1966some, luo2024lowrank}.However, these methods remain largely driven by sparse observation fitting or learned priors. Continuous representability alone does not ensure physical consistency where observations are absent.
\endgroup

\begingroup
For systems with an available equation constraint, physical laws provide an additional source of supervision where sparse data provide none~\cite{xu2023pinns}. The observation loss is evaluated only at sensor locations, whereas the desired reconstruction is defined over a continuous domain. Finite-difference equation residuals can be imposed at off-sensor collocation points, providing an additional inductive bias in regions where no measurements are available~\cite{REN2022114399}. Thus, PDE residuals are not merely post-hoc physical-consistency metrics or generic smoothing penalties; they reduce the ambiguity of sparse reconstruction by constraining the decoded field in the observation null space. To use such regularization, the representation must be continuously evaluable at sensors, collocation points, and neighboring stencil locations, while remaining compact enough for sparse inverse recovery. This motivates a structured continuous representation that separates spatial coordinate dependence from low-dimensional time-varying coefficients.
\endgroup

\begingroup
Motivated by this view, we propose \textbf{PhysFTM}, which represents the continuous dynamics as
\[
\hat u(\mathbf{x},t,c)=\mathcal D_{\Phi}(G(t))(\mathbf{x},c),
\]
where \(\mathcal D_{\Phi}\) is a spatially continuous Functional Tucker decoder and \(G(t)\) is a continuous trajectory in Tucker-core space. The decoder can be evaluated at sensor locations, dense grids, and collocation points, while temporal evolution is modeled through low-dimensional core coefficients. To make the spatial decoder reliable under sparse supervision, PhysFTM parameterizes its mode-wise factors in reproducing kernel Hilbert spaces (RKHSs) and regularizes their RKHS norms~\citep{aronszajn1950theory, wahba1990spline}. This replaces unconstrained neural mode functions with explicitly regularized functional factors. Together with finite-difference PDE residuals at collocation points, the resulting RKHS-FTM decoder provides both continuous spatial evaluability and physics-guided off-sensor guidance.
\endgroup

\begingroup
Our contributions are summarized as follows.
1) We propose PhysFTM, a functional Tucker framework that combines RKHS-based spatial factors with a continuous trajectory in Tucker-core space for sparse spatiotemporal reconstruction, paving the way for inquiries of the field at arbitrary times and positions.
2) We introduce a hybrid solve-learn optimization scheme that solves conditionally linear Tucker cores with regularized conjugate gradient method while learning RKHS factors under observation, RKHS, and PDE losses.
3) We provide theoretical guarantees for reconstruction via the proposed PhysFTM framework, and evaluate PhysFTM on Allen--Cahn, Navier--Stokes, Heat equations, and a gridded Copernicus ocean reanalysis, covering 2D/3D reconstruction, showing improved accuracy,off-sensor reconstruction accuracy, physical consistency, and missing frame reconstruction.
\endgroup

%% file: sections/02_related_work.tex
\section{Related Work}

\paragraph{Sparse reconstruction of physical fields.}
\begingroup
 Early deep-learning approaches often transformed sparse measurements into structured grid representations, e.g., through Voronoi-assisted representations, followed by convolutional decoders for full-field recovery~\citep{fukami2021global}. Senseiver~\citep{santos2023senseiver} encodes unordered sparse sensor sets with attention and decodes field values at query coordinates.
SHRED and SINDy-SHRED~\citep{williams2024shred,gao2026sindyshred} reconstruct high-dimensional spatiotemporal states from sensor histories using recurrent latent representations or identified latent dynamics.
\endgroup
STRIDE~\citep{stride2026} combines temporal encoders with modulated implicit decoders for continuous field reconstruction. Additionally, generative approaches were explored for physical field recovery, including conditional neural-field latent diffusion for spatiotemporal turbulence~\citep{du2024confield} and diffusion in Functional Tucker space for full-field dynamics from irregular sparse observations~\citep{chen2025sdift}. Existing approaches often rely on neural decoders that remain weakly constrained in off-sensor regions under sparse observations.

\paragraph{Functional tensor and low-rank continuous representations.}
Low-rank tensor decompositions provide compact multilinear representations for high-dimensional data, with Tucker decomposition being one of the classical formulations based on mode factors and a core tensor~\citep{tucker1966some,kolda2009tensor,gao2025lowrank}. To extend such representations beyond fixed grids, one may consider continuous or functional variants that replace discrete factor matrices with coordinate-dependent factor functions, enabling field reconstruction and evaluation at arbitrary locations; see, e.g., \citep{luo2024lowrank,fang2024funbat}.
Functional Tucker models, where observations are reconstructed by shared latent functions and time-dependent Tucker cores~\citep{chen2025sdift}, were adapted as latent spaces for sparse physical dynamics. Related functional tensor approaches include~\citep{dolgov2021functional,wang2022nonparametric,wang2023demote,steidle2026adaptive}, which offer alternative representations and priors for continuous tensor modeling. %

\paragraph{Physics-informed reconstruction and PDE supervision.}
Physics-informed learning incorporates governing equations into neural approximation by penalizing PDE residuals at collocation points. PINNs are the canonical example, using residual losses together with data, boundary, or initial-condition terms to solve forward and inverse PDE problems~\citep{raissi2019physics}; broader physics-informed machine learning has been reviewed as a way to combine data with scientific constraints~\citep{karniadakis2021physics}. Physics-informed neural operators extend this idea to operator learning by combining data losses with PDE residuals or physical consistency terms~\citep{li2024physics,yang2025spatially}. Discrete and spatiotemporal architectures such as PhyCRNet impose residual constraints in convolutional-recurrent models for PDE dynamics~\citep{REN2022114399}. In reconstruction settings, Hidden Fluid Mechanics uses Navier--Stokes constraints to infer velocity and pressure fields from indirect observations~\citep{raissi2020hidden}, and physics-informed diffusion models incorporate physical conditions to improve high-fidelity flow-field reconstruction~\citep{shu2023physics}. These works show that PDE residuals can provide supervision beyond labeled samples and improve physical consistency.

%% file: sections/03_methodology.tex
\section{Methodology of PhysFTM}
\label{sec:method}

\begin{figure}[t]
  \centering
  \includegraphics[width=\linewidth]{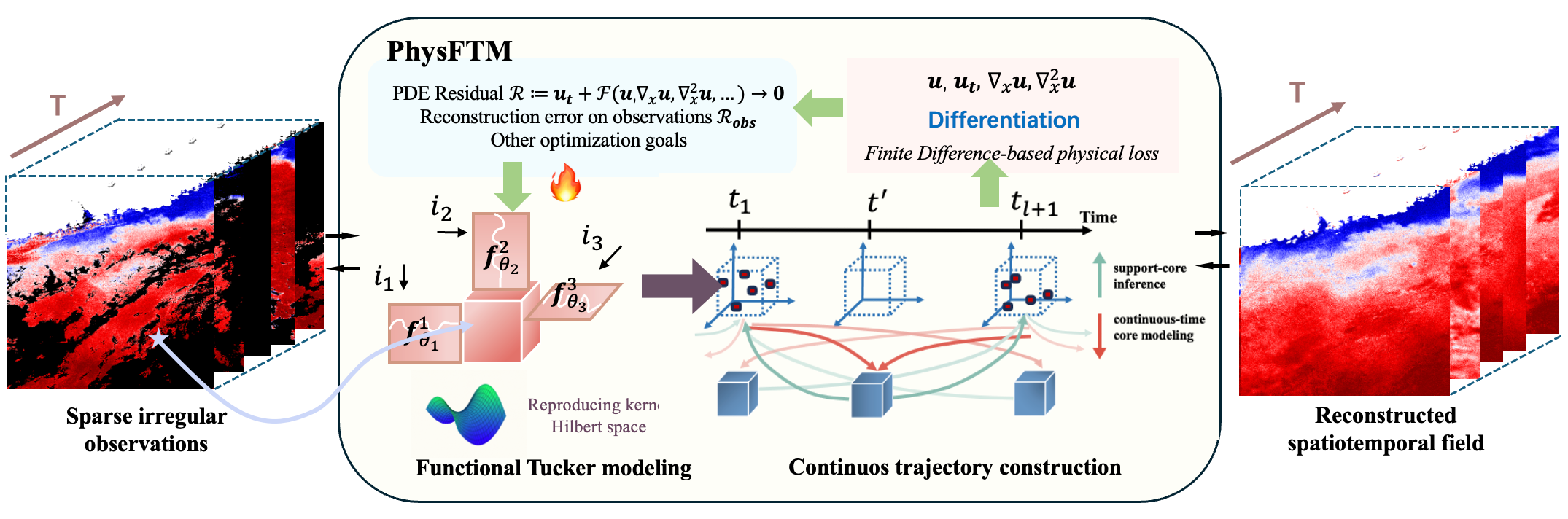}
  \caption{Overall framework of PhysFTM. PhysFTM reconstructs  sparse grid-based sensor observations by learning an RKHS-regularized Functional Tucker spatial decoder constrained by observation loss and finite-difference PDE residuals. The inferred Tucker cores are then modeled as a continuous trajectory in core space and decoded back to the physical domain to obtain the reconstructed spatiotemporal field.}
  \label{fig:framework}
\end{figure}

\subsection{Problem setup and PhysFTM decomposition}
\label{sec:formulation}

\paragraph{Sparse observations.}
\begingroup
For clarity, we describe the formulation for a single trajectory
in two spatial dimensions and omit the trajectory index.
Let $u(\mathbf{x},t,c)$ denote the $c$-th physical channel at
time $t$ and spatial coordinate $\mathbf{x}=(x,y)$.
The data are represented by
$Y\in\mathbb{R}^{H\times W\times T\times C}$,
where $H\times W$ is the spatial grid resolution,
$T$ is the number of frames, and $C$ is the number of physical
channels. Only a sparse subset of entries is available as
observations. We denote the observation mask by
$M\in\{0,1\}^{H\times W\times T\times C}$
and the observed index set by
\[
\Omega=\{(h,w,t,c):M_{h,w,t,c}=1\}.
\]
All trajectories share a common nominal time grid.
Sensor locations are fixed over time within each trajectory,
and other spatial locations are referred to as
\emph{off-sensor locations}.
Frames with sensor observations are referred to as
\emph{support frames}, while frames without sensor observations
are referred to as \emph{missing frames}.
We aim to reconstruct the full field
$\hat u(\mathbf{x},t,c)$ on both support and missing frames
from these sparse observations.
\endgroup

\paragraph{Spatial--temporal factorization.}
\begingroup
PhysFTM combines a spatially continuous Functional Tucker
decoder with a time-dependent Tucker core:
\begin{equation}
  \hat u(\mathbf{x},t,c)
  = \mathcal{D}_{\Phi}\big(G(t)\big)(\mathbf{x},c),
  \label{eq:physftm-composition}
\end{equation}
where $\mathcal{D}_{\Phi}$ is a spatial decoder defined by
RKHS-parameterized Functional Tucker factors, and $G(t)$ is
the time-dependent Tucker core.
Sparse sensor observations in support frames are used to
infer the corresponding cores $G(t_j)$.
A continuous-time model then estimates the cores of missing
frames from these inferred support-frame cores, and
$\mathcal{D}_{\Phi}$ decodes them into reconstructed fields.
The spatial decoder provides a continuous spatial
representation and can be regularized by PDE residuals,
while temporal modeling operates in the compact Tucker-core
space.
\endgroup

\subsection{Physics-informed RKHS Functional Tucker decoder}
\label{sec:rkhs-ftm-decoder}

\paragraph{RKHS mode factors and Tucker decoding.}
For each active spatial mode $d\in\{x,y\}$, PhysFTM defines $R_d$ factor functions $\{\phi_{d,r}\}_{r=1}^{R_d}$. Each factor belongs to a reproducing kernel Hilbert space induced by a one-dimensional positive definite kernel $k_d$ and is represented by a finite kernel expansion
\begin{equation}
  \phi_{d,r}(\xi)=\sum_{m=1}^{M_d}a_{d,m,r}\,k_d(\xi,z_{d,m}),
  \label{eq:rkhs-factor}
\end{equation}
where $\{z_{d,m}\}_{m=1}^{M_d}$ are inducing points and $A_d=[a_{d,m,r}]\in\mathbb{R}^{M_d\times R_d}$ are learnable RKHS coefficients. By using Eq.~\eqref{eq:rkhs-factor}, we provide the following spatial decoders via Tucker decomposition. For a scalar field at a support frame $t$, the spatial decoder is
\begin{equation}
  \hat u(t,x,y)
  =
  \sum_{p=1}^{R_x}\sum_{q=1}^{R_y}
  G_{p,q,t}\,\phi_{x,p}(x)\phi_{y,q}(y),
  \label{eq:scalar-decode}
\end{equation}
where $G$ is a core tensor. For multichannel fields, the spatial decoder is modeled by
\begin{equation}
  \hat u(t,x,y,c)
  =
  \sum_{p=1}^{R_x}\sum_{q=1}^{R_y}
  G_{p,q,t,c}\,\phi_{x,p}(x)\phi_{y,q}(y).
  \label{eq:multichannel-decode}
\end{equation}
Note that factors $\phi_{x,p}(x)$ and $\phi_{y,q}(y)$ are shared between channels, and the channel index is stored in the core tensor. The decoder can be evaluated at observed sensors, dense grid points, and off-grid PDE collocation points.

\paragraph{RKHS and physics-informed loss functions and alternating-minimization strategy.}
\label{sec:hybrid-optimization}
\begingroup
PhysFTM exploits the block structure of Eq.~\eqref{eq:physftm-composition}.In general, we propose an alternating-minimization strategy to optimize the factors and core tensor. Specifically, with fixed RKHS factors, support-time Tucker cores can be inferred by solving a least-squares problem. For RKHS factors, since the factors are into 143
the PDE residuals, they are learned by gradient descent under observation, RKHS, and PDE losses. %
\endgroup

For a mini-batch and fixed factors, the support-time core sequence is obtained by
\begin{equation}
  \{G_t^\star\}_{t=1}^{T_s}
  =
  \arg\min_{\{G_t\}}
  \sum_{t=1}^{T_s}
  \big\|M_t\odot(\Phi\operatorname{vec}(G_t)-Y_t)\big\|_2^2
  +\lambda_{\mathrm{ridge}}\sum_{t=1}^{T_s}\|G_t\|_F^2
  +\beta_{\mathrm{TV}}\sum_{t=2}^{T_s}\|G_t-G_{t-1}\|_F^2.
  \label{eq:core-solve}
\end{equation}
\begingroup
Here $T_s$ denotes the number of support frames, $\Phi(\mathbf{x})=\bigotimes_{d=1}^{d_s}\phi_d(x_d)$ is the tensor-product spatial feature, and $M_t$ is the sparse observation mask at time $t$. The subproblem is solved by conjugate gradients using implicit matrix-vector products. Note that the matrix $\Phi(\mathbf{x})$ is never explicitly computed.
\endgroup

\begingroup
After solving the cores, the RKHS basis coefficients are
updated by AdamW by minimizing
\begin{equation}
  \mathcal{L}_{\mathrm{basis}}
  =
  \mathcal{L}_{\mathrm{obs}}
  +\lambda_{\mathrm{RKHS}}\mathcal{L}_{\mathrm{RKHS}}
  +\lambda_{\mathrm{PDE}}\mathcal{L}_{\mathrm{PDE}}
  +\lambda_t\mathcal{L}_{\mathrm{time}},
  \label{eq:basis-total}
\end{equation}
with the loss terms defined below. The observation loss is
\begin{equation}
  \mathcal{L}_{\mathrm{obs}}
  =
  \frac{1}{|\Omega|}
  \sum_{(t,h,w,c)\in\Omega}
  \big(\hat Y_{t,h,w,c}-Y_{t,h,w,c}\big)^2
\end{equation}
This is the RMSE over observed scalar entries of the
channelwise normalized fields. For fixed factors, the core
update uses the squared-error subproblem in
Eq.~\eqref{eq:core-solve}.
Note that in practice, for a mini-batch of $N_b$ trajectories, let $\Omega_b$ denote
the observed index set of trajectory $b$, we evaluate the loss on each $\Omega_b$ separately and compute the weighted average by Eq.~\eqref{eq:core-solve}.

The RKHS regularization term controls the complexity of the
spatial factor functions:
\begin{equation}
  \mathcal{L}_{\mathrm{RKHS}}
  =
  \sum_{d\in\{x,y\}}
  \mathrm{tr}(A_d^\top K_dA_d),
  \label{eq:rkhs-reg}
\end{equation}
where $K_d$ is the kernel Gram matrix over inducing points of mode $d\in\{x,y\}$.
Given a mini-batch of trajectories and spatial collocation points $\mathcal{Q}$, the decoder evaluates $\hat u(\mathbf{x}, t, c)$ for each $\mathbf{x}\in\mathcal{Q}$. For a dataset-specific finite-difference residual operator $\mathcal{R}_h$, the PDE loss is
\begin{equation}
  \mathcal{L}_{\mathrm{PDE}}
  =
  \frac{1}{N_b|\mathcal{Q}|T'}
  \sum_{b=1}^{N_b}
  \sum_{\mathbf{x}\in\mathcal{Q}}
  \sum_{t\in\mathcal{T}'}
  \big\|\mathcal{R}_h[\hat u_b](\mathbf{x},t)\big\|_2^2,
  \label{eq:pde-loss}
\end{equation}
where $\mathcal{T}'$ contains the timestamps of support
frames with valid finite-difference stencils, and
$T'=|\mathcal{T}'|$.
The derivative stencils and valid spatial and temporal
points depend on the dataset; see
Appendix~\ref{app:pdeloss}.

The time-kernel term regularizes the inferred
support-frame core sequence:
\begin{equation}
  \mathcal{L}_{\mathrm{time}}
  =
  \frac{1}{D_G}
  \sum_{b=1}^{N_b}
  \mathrm{tr}\left(
    (G_b^\flat)^\top K_t^{-1}G_b^\flat
  \right),
  \label{eq:time-gp}
\end{equation}
where $G_b^\flat\in\mathbb{R}^{T_s\times D_G}$ is the
flattened core sequence of trajectory $b$,
$T_s$ is the number of support frames,
$D_G=CR_xR_y$ is the flattened-core dimension, and
$K_t$ is an RBF kernel matrix over the timestamps of
support frames.
For three-dimensional fields, the RKHS regularization
also includes the $z$-mode factors, and
$D_G=CR_xR_yR_z$.
\endgroup

\paragraph{RKHS-Tucker reconstruction analysis.}
We provide the reconstruction error analysis of the RKHS-Tucker formulation in Appendix~\ref{app:Theory}.
\begin{theorem}[Fixed-rank sparse reconstruction]
\label{prop:fixed-rank-reconstruction}
Assume that 1) the observation points are i.i.d. samples from $\rho$; 2) the noise is conditionally sub-Gaussian; 3) the tensor RKHS integral operator admits a product spectral decomposition; 4) the target field satisfies the source condition $u=T^v g$ with $v>1/2$; 5) and the mode-wise eigenvalues satisfy $\lambda_{j,m}\leq K_0 m^{-\beta}$ with $\beta(2v-1)>1$.
Then, for a rank-$r$ RKHS-Tucker estimator $\hat u_r$, with probability at least $1-\delta$,
\begin{equation}
  \|\hat u_r-u\|_{L^2(\rho)}
  \lesssim
  C_A r^{-s}
  +
  C_S
  \sqrt{
  \frac{r^d+\log(1/\delta)}{n}
  },
  \qquad
  s=\frac{\beta(2v-1)-1}{2}.
  \label{eq:fixed-rank-reconstruction-bound}
\end{equation}
\end{theorem}

Note that the first term in Eq.~\eqref{eq:fixed-rank-reconstruction-bound} is the truncation error induced by the finite Tucker rank, and the second term comes from the statistical error induced by sparse sampling.

\begin{corollary}[Sample complexity]
\label{cor:balanced-reconstruction-rate}
Let the truncation rank be $r_\star\asymp n^{1/(2s+d)}$, which balances approximation and statistical errors.
Then, under the conditions of Theorem~\ref{prop:fixed-rank-reconstruction},
if the sample size satisfies
\begin{equation}
  n
  \gtrsim
  \left(\frac{C}{\epsilon}\right)^{
  \frac{
  2\left(\beta(2v-1)-1+d\right)
  }{
  \beta(2v-1)-1
  }
  }
  +
  \frac{C^2\log(1/\delta)}{\epsilon^2},
\end{equation}
the $L^2$-norm $\|\hat u_{r_\star}-u\|_{L^2(\rho)}$ is smaller than $\epsilon$ with probability at least $1-\delta$.
\end{corollary}

\begingroup\subsection{Continuous-time core trajectory and missing frames reconstruction}\endgroup
\label{sec:continuous-core-trajectory}

After learning the spatial RKHS-FTM decoder, PhysFTM freezes the RKHS factor functions and represents temporal variation only through the time-dependent Tucker core.
\begingroup
The Tucker core provides a compact coefficient representation for modeling temporal dynamics under the spatial decoder. We derive the temporal model from support-core estimates inferred with the RKHS decoder learned during the physics-constrained spatial stage.
\endgroup

\paragraph{Support-core inference with a mean-core reference.}
For a new trajectory, the learned RKHS factors are frozen and reused as a deterministic spatial decoder.
The remaining task is to infersupport-frame cores from sparse sensor measurements.
Under severe sparsity, this inverse problem is highly underdetermined: the sensing operator observes only a small subset of entries, leaving a large null space in core space. If the core is solved directly with a ridge penalty, the solution can be biased toward a minimum-norm core; in practice, especially with finite-iteration CG and extremely sparse masks, this often leads to near-zero cores and low-energy reconstructions.

Therefore, we introduce a residual correction:
\begin{equation}
    G_t = G_t^{\mathrm{ref}}+\Delta G_t
\end{equation}
on the data-driven mean-core reference $G_t^{\mathrm{ref}}$ computed from the training set,
\begin{equation}
  G_t^{\mathrm{ref}}
  =
  \frac{1}{N_{\mathrm{train}}}
  \sum_{i=1}^{N_{\mathrm{train}}}G_{i,t}^{\mathrm{train}}.
\end{equation}
Instead of regularizing the absolute core, we penalize only the correction.
The core-correction subproblem is defined on the complete native time grid:
\begin{equation}
\begin{aligned}
\Delta G^\star=\arg\min_{\{\Delta G_t\}_{t=1}^{T}}\quad
&\sum_{t=1}^{T}\big\|M_t\odot\big(\Phi\operatorname{vec}(G_t^{\mathrm{ref}}+\Delta G_t)-Y_t\big)\big\|_2^2\\
&+\lambda_{\mathrm{supp}}\sum_{t=1}^{T}\|\Delta G_t\|_F^2\\
&+\beta_{\mathrm{supp}}\sum_{t=2}^{T}\|\Delta G_t-\Delta G_{t-1}\|_F^2.
\end{aligned}
\end{equation}
Here $T$ is the number of native timestamps. The observation mask is zero at missing frames, while the last summand couples corrections at adjacent native timestamps.
The inferred support-time cores are
\begin{equation}
    \bar G_t = G_t^{\mathrm{ref}}+\Delta G_t^\star.
\end{equation}
This residual formulation keeps the solution close to a plausible training-set core trajectory when observations are insufficient, while still allowing sparse measurements to correct the trajectory through \(\Delta G\).

\begingroup
\paragraph{Temporal core modeling.}
Let $\{(t_j,\bar G(t_j))\}_{j=1}^{T_s}$ denote the inferred
support-frame cores and their timestamps.
A continuous-time model $H_\psi$ estimates the cores of
missing frames:
\begin{equation}
  \hat G(\tau)
  =
  H_\psi\big(
    \{(t_j,\bar G(t_j))\}_{j=1}^{T_s},\tau
  \big),
  \qquad \tau\in\mathcal{T}_{\mathrm{miss}},
  \label{eq:continuous-core-head}
\end{equation}
where $\mathcal{T}_{\mathrm{miss}}$ contains the timestamps
of missing frames within $[t_1,t_{T_s}]$.

We implement $H_\psi$ as a learned residual correction
to cubic Hermite interpolation in core space.
Let $B(\tau)$ denote the Hermite interpolant through
the support-frame cores.
Between two adjacent support-frame timestamps $t_a$ and $t_b$,
the estimated core is
\begin{equation}
  \hat G(\tau)
  =
  B(\tau)+\gamma(\tau)\Delta_\psi(\tau),
  \qquad
  \gamma(\tau)
  =
  \sin^2\left(
    \pi\frac{\tau-t_a}{t_b-t_a}
  \right),
  \label{eq:gated-residual-core}
\end{equation}
where $\Delta_\psi(\tau)$ is generated by a lightweight
ODE-RNN in flattened core space.
The gate is zero at both endpoints, preserving the
inferred support-frame cores.

\paragraph{Missing-frame decoding.}
The reconstructed field at spatial location $\mathbf{x}$
and missing-frame timestamp $\tau$ is obtained through
the learned spatial decoder:
\begin{equation}
  \hat u(\mathbf{x},\tau,c)
  =
  \mathcal{D}_{\Phi}\big(\hat G(\tau)\big)(\mathbf{x},c).
  \label{eq:final-query-decode}
\end{equation}
\endgroup

%% file: sections/04_experiments_neurips.tex
\section{Experiments}
\label{sec:experiments}
\vspace{-0.55em}

\subsection{Experimental setup}
\label{sec:exp-setup}
\begingroup
\vspace{-0.45em}

\noindent\textbf{Datasets.}
We use \textbf{Allen--Cahn (AC2D)}, a scalar reaction–diffusion benchmark with 200 trajectories of 201 frames on a $256\times256$ grid, split into 160/20/20 train/validation/test trajectories; \textbf{Navier--Stokes (NS2D)}, an incompressible flow benchmark with $(u,v,\omega)$ channels and 800 trajectories of 201 frames on a $128\times128$ grid, split into 640/80/80; \textbf{Heat3D}, a scalar three-dimensional heat-diffusion
benchmark with 200 trajectories of 50 frames on a $64^3$ grid,
split into 160/20/20 train/validation/test trajectories;
and \textbf{Ocean3D}, a three-dimensional ocean-velocity
reanalysis dataset derived from the Copernicus Marine
GLORYS12V1 product~\citep{cmems_glorys12}, with $(u,v)$ channels
and 100 sequences of 90 frames on a $32\times64\times64$ grid,
split chronologically into 80/10/10
train/validation/test sequences.
Dataset details are in Appendix~\ref{app:dataset}.

\noindent\textbf{Observation and missing-frame protocols.}
During training, $15\%$ of spatial locations are sampled to simulate sparse grid-based sensors. The sensor mask is fixed over time within each trajectory and shared by physical channels, while different trajectories use different masks. We refer to the sensor positions of each trajectory as sensor locations and all remaining spatial locations as off-sensor locations. We report spatial sensor ratios $s\in\{0.01,0.03,0.15\}$. We use two missing-frame protocols: random-drop-half independently retains internal frames as support frames with probability $0.5$ and always retains both endpoints, whereas temporal-interval-missing withholds sensor observations from a contiguous internal block of $20\%$ of the frames. Frame masks are trajectory specific; full protocol and implementation details are in Appendix~\ref{app:native-time-protocols}.

\noindent\textbf{Compared Methods.}
We compare PhysFTM with SDIFT~\citep{chen2025sdift}, LRTFR~\citep{luo2024lowrank}, SIREN-MLP~\citep{sitzmann2020siren}, PINNs~\citep{raissi2019physics}, and RecFNO~\citep{zhao2024recfno}. Their conditioning, training, and test-time protocols are summarized in Appendix~\ref{app:comparison-protocols}.

\noindent\textbf{Metrics.}
We report root mean squared error (RMSE) at sensor locations in support frames (S-sup), off-sensor locations in support frames (Off-sup), sensor locations in missing frames (S-miss), and off-sensor locations in missing frames (Off-miss). The four disjoint subsets and aggregation rules are defined in Appendix~\ref{app:fourway-subsets}. All metrics use inverse-normalized physical variables.
\endgroup

\subsection{Reconstruction results}
\label{sec:exp-results}
\begingroup
\vspace{-0.45em}

\noindent\textbf{Overall reconstruction under missing-frame protocols.}
We first compare reconstruction at $s=0.03$ across the four datasets and both missing-frame protocols. Table~\ref{tab:end-to-end-cvprlike} reports metrics separately for support frames and missing frames, each evaluated at sensor locations and off-sensor locations.

\begin{table}[!tbp]
\caption{Reconstruction RMSE on four datasets at spatial sensor ratio $s=0.03$. The two column groups report the four disjoint evaluation subsets, defined by sensor/off-sensor locations and support/missing frames, under random-drop-half and temporal-interval-missing. Lower is better. For Heat3D and Ocean3D, the PhysFTM interval-missing results use the random-drop-half-trained temporal checkpoint. (\textbf{Bold}: best, \underline{Underline}: second best)}
\label{tab:end-to-end-cvprlike}
\centering
\small                          %
\setlength{\tabcolsep}{4pt}   %
\begin{tabular}{@{}lrrrrrrrr@{}}
\toprule
\multirow{2}{*}{Method}
& \multicolumn{4}{c}{Random-drop-half}
& \multicolumn{4}{c}{Temporal-interval-missing}\\
\cmidrule(lr){2-5}\cmidrule(lr){6-9}
& S-sup & Off-sup & S-miss & Off-miss & S-sup & Off-sup & S-miss & Off-miss\\
\midrule
\multicolumn{9}{c}{\textit{AC2D}}\\
SDIFT     & 0.35804 & 0.38080 & 0.38030 & 0.38151 & 0.35699 & 0.36199 & 0.39071 & 0.39259\\
LRTFR     & 0.31170 & \underline{0.32220} & 0.31340 & \underline{0.32038} & 0.31006 & \underline{0.31676} & 0.32059 & \underline{0.32781}\\
PhysFTM   & \textbf{0.01735} & \textbf{0.08710} & \underline{0.05035} & \textbf{0.09988} & \textbf{0.01633} & \textbf{0.09397} & \underline{0.03953} & \textbf{0.07924}\\
RecFNO    & \underline{0.01848} & 0.45283 & \textbf{0.02178} & 0.45523 & \underline{0.01811} & 0.45875 & \textbf{0.02116} & 0.43680\\
SIREN-MLP & 0.34868 & 0.34932 & 0.35013 & 0.35076 & 0.34605 & 0.34664 & 0.35819 & 0.35898\\
PINNs     & 0.35166 & 0.35243 & 0.35268 & 0.35331 & 0.34874 & 0.34943 & 0.36508 & 0.36622\\
\midrule
\multicolumn{9}{c}{\textit{NS2D}}\\
SDIFT     & 0.52867 & 0.50970 & 0.51003 & 0.51027 & 0.53019 & 0.53052 & 0.50434 & 0.50275\\
LRTFR     & 0.13245 & 0.14720 & 0.13447 & 0.14576 & 0.13515 & 0.14682 & \underline{0.19025} & \underline{0.19661}\\
PhysFTM   & \textbf{0.00669} & \textbf{0.06210} & \textbf{0.03240} & \textbf{0.06498} & \textbf{0.00435} & \textbf{0.06282} & \textbf{0.14949} & \textbf{0.15056}\\
RecFNO    & \underline{0.06810} & \underline{0.09757} & \underline{0.07820} & \underline{0.10441} & \underline{0.06906} & \underline{0.09910} & 0.35803 & 0.35943\\
SIREN-MLP & 0.24555 & 0.24526 & 0.24465 & 0.24441 & 0.24797 & 0.24777 & 0.21369 & 0.21354\\
PINNs     & 0.34254 & 0.34177 & 0.34101 & 0.34055 & 0.34722 & 0.34657 & 0.31892 & 0.31848\\
\midrule
\multicolumn{9}{c}{\textit{Heat3D}}\\
SDIFT     & 0.05811 & 0.05808 & 0.06303 & 0.06272 & 0.06284 & 0.06268 & 0.05016 & 0.04989\\
LRTFR     & \underline{0.01212} & \underline{0.01222} & \underline{0.01309} & \underline{0.01315} & \underline{0.01210} & \underline{0.01225} & \underline{0.00577} & \underline{0.00577}\\
PhysFTM   & \textbf{0.00108} & \textbf{0.00125} & \textbf{0.00575} & \textbf{0.00574} & \textbf{0.00122} & \textbf{0.00137} & \textbf{0.00310} & \textbf{0.00305}\\
RecFNO    & 0.28736 & 0.77847 & 0.26372 & 0.72034 & 0.28629 & 0.77526 & 0.22597 & 0.64187\\
SIREN-MLP & 0.04080 & 0.04087 & 0.04211 & 0.04205 & 0.04342 & 0.04344 & 0.02749 & 0.02742\\
PINNs     & 0.04000 & 0.04011 & 0.04271 & 0.04264 & 0.04314 & 0.04319 & 0.03353 & 0.03334\\
\midrule
\multicolumn{9}{c}{\textit{Ocean3D}}\\
SDIFT     & 0.27863 & 0.28065 & 0.27585 & 0.27838 & 0.27494 & 0.27725 & 0.28712 & 0.28941\\
LRTFR     & \underline{0.11038} & \underline{0.11450} & \underline{0.11767} & \underline{0.12155} & \underline{0.10506} & \underline{0.10829} & \textbf{0.18062} & \textbf{0.18377}\\
PhysFTM   & \textbf{0.00632} & \textbf{0.06574} & \textbf{0.07133} & \textbf{0.08858} & \textbf{0.00595} & \textbf{0.06382} & \underline{0.18985} & \underline{0.18565}\\
RecFNO    & 0.73827 & 0.97074 & 0.77688 & 0.96302 & 0.71841 & 0.94379 & 0.99191 & 1.05383\\
SIREN-MLP & 0.20158 & 0.20168 & 0.20024 & 0.20094 & 0.19773 & 0.19833 & 0.21199 & 0.21151\\
PINNs     & 0.30163 & 0.30100 & 0.30232 & 0.30194 & 0.29765 & 0.29739 & 0.31870 & 0.31725\\
\bottomrule
\end{tabular}%
\end{table}

PhysFTM obtains the lowest RMSE in most reported dataset--protocol--subset comparisons. The gains are especially clear at off-sensor locations in support frames on AC2D, NS2D, and Heat3D. These results support the effectiveness of the proposed representation for fixed-sensor reconstruction with missing frames across both 2D and 3D grids.

\noindent\textbf{Complementary physical diagnostics.}
Table~\ref{tab:ns-physics} reports NS2D missing-frame accuracy together with divergence, curl-derived vorticity, and kinetic-energy-spectrum errors. The diagnostics are computed on inverse-normalized full grids of missing frames; their formulas and cross-frame reducers are given in Appendix~\ref{app:physical-diagnostics}.

\begin{table}[!tbp]

\caption{NS2D missing-frame reconstruction and complementary physical diagnostics at $s=0.03$. Divergence, vorticity, and spectrum errors are evaluated on full grids of missing frames; Off-miss is evaluated at off-sensor locations in missing frames. Lower is better.}
\label{tab:ns-physics}
\centering
\small
\setlength{\tabcolsep}{4pt}
\begin{tabular}{@{}llrrrr@{}}
\toprule
Protocol & Method & Off-miss & Divergence & Vort. RMSE & Spectrum \\
\midrule
\multirow{6}{*}{Random}
& SDIFT     & 0.510274 & 0.352407 & 0.916343 & 0.877309\\
& LRTFR     & 0.145757 & 0.459044 & 0.237923 & \underline{0.049954}\\
& PhysFTM   & \textbf{0.064976} & \underline{0.134170} & \textbf{0.109304} & \textbf{0.033599}\\
& RecFNO    & \underline{0.104413} & 0.459014 & \underline{0.118608} & 0.067430\\
& SIREN-MLP & 0.244405 & 1.887604 & 0.623477 & 0.408029\\
& PINNs     & 0.340554 & \textbf{0.120243} & 0.586433 & 0.398297\\
\midrule
\multirow{6}{*}{Interval}
& SDIFT     & 0.502748 & 0.344546 & 0.931635 & 0.892291\\
& LRTFR     & \underline{0.196613} & 0.616958 & 0.272195 & \underline{0.055034}\\
& PhysFTM   & \textbf{0.150558} & \underline{0.129873} & \textbf{0.150436} & \textbf{0.036346}\\
& RecFNO    & 0.359434 & 0.342736 & \underline{0.211528} & 0.055423\\
& SIREN-MLP & 0.213537 & 1.863564 & 0.546894 & 0.368662\\
& PINNs     & 0.318479 & \textbf{0.120243} & 0.586433 & 0.349142\\
\bottomrule
\end{tabular}
\end{table}

PhysFTM has the lowest Off-miss, vorticity, and spectrum errors under both protocols, while PINNs has the lowest divergence residual. The result indicates a favorable accuracy--consistency trade-off. Complete three-ratio diagnostics are reported in Appendix~\ref{app:physical-diagnostics}.

\begin{figure}[!tbp]

  \centering
  \includegraphics[width=0.98\linewidth]{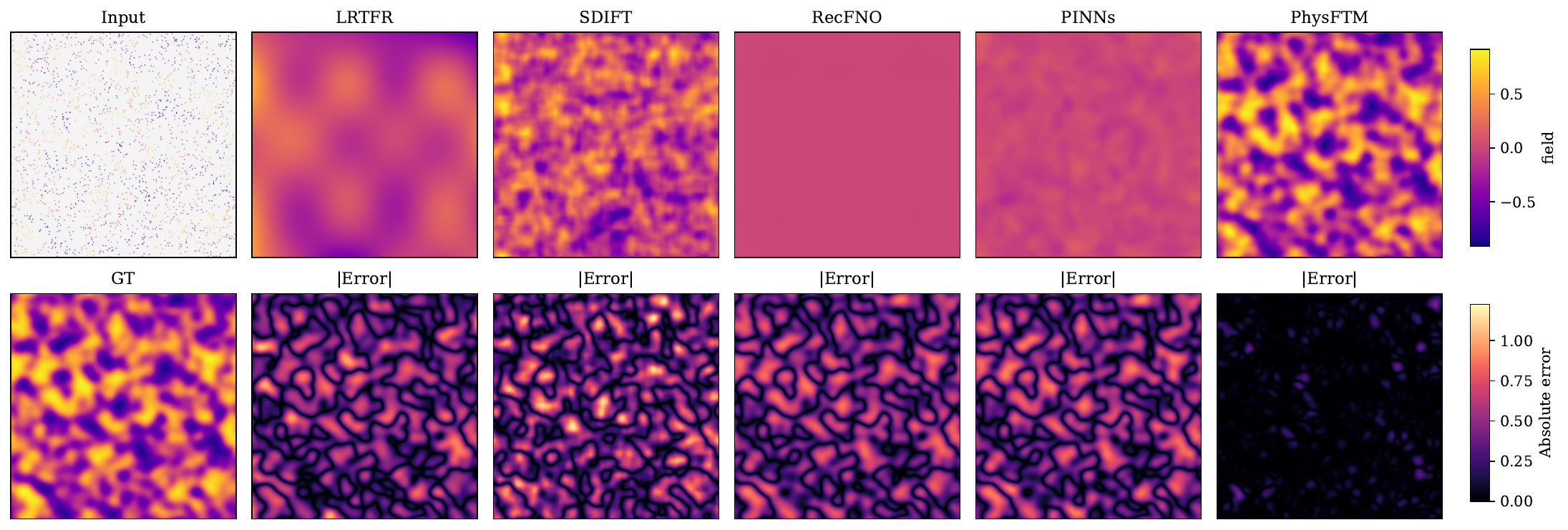}
  \vspace{0.5em}
  \includegraphics[width=0.98\linewidth]{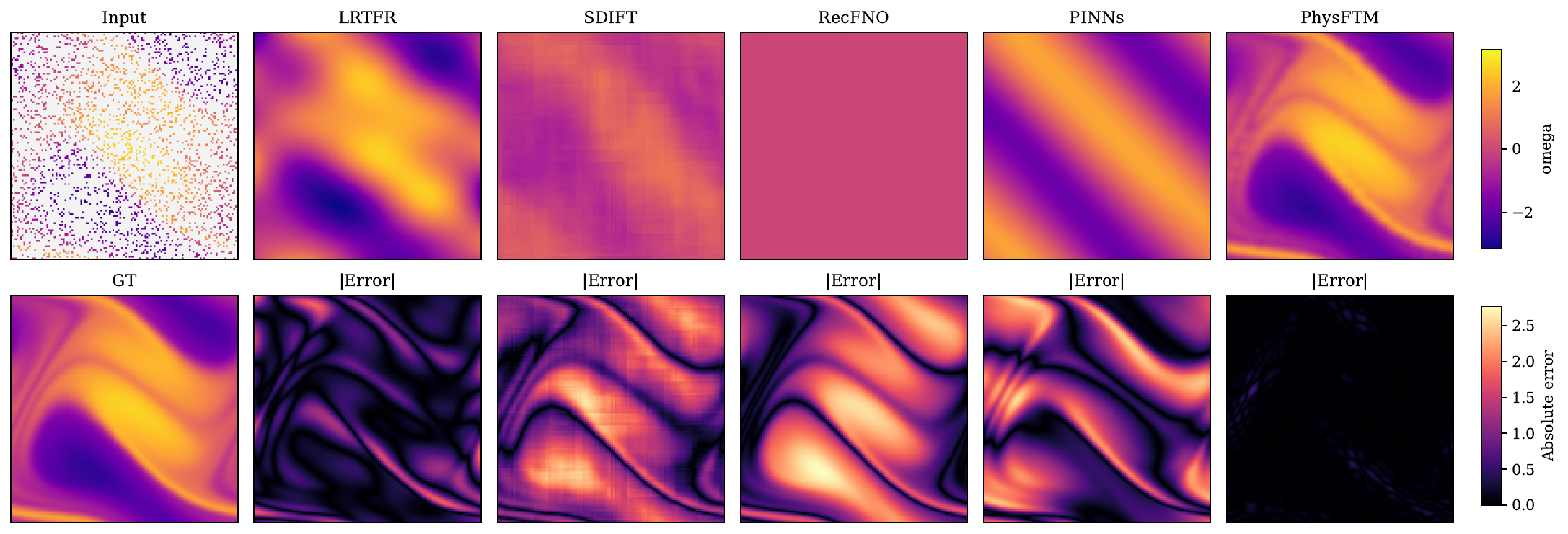}
  \caption{Qualitative comparisons on AC2D (top) at $s=0.03$ and the NS2D vorticity channel (bottom) at $s=0.15$. The upper row of each block shows the sparse input and predictions from LRTFR, SDIFT, RecFNO, PINNs, and PhysFTM; the lower row shows the ground truth and the corresponding absolute-error maps.}
  \label{fig:qual-cvprlike}
  \vspace{-0.8em}
\end{figure}
\endgroup

\subsection{Model analysis}
\label{sec:model-analysis}
\begingroup
\vspace{-0.45em}

\noindent\textbf{The effects of individual components and block-update strategies are evaluated through ablations.}
Table~\ref{tab:field-collocation-cvprlike} compares RKHS regularization, PDE regularization, and core/factor block-update strategies with their corresponding effective objectives, without temporal missingness.

\begin{table}[!tbp]

\caption{Component and block-update-strategy ablation on AC2D and NS2D at $s=0.03$, without temporal missingness. RMSEs use all evaluated frames; PDE is evaluated at disjoint held-out off-sensor points and centered-stencil-valid internal times.}
\label{tab:field-collocation-cvprlike}
\centering
\small
\setlength{\tabcolsep}{4pt}
\begin{tabular}{@{}llrrrr@{}}
\toprule
Data & Variant & PDE & Full & Sensor & Off-sensor\\
\midrule
\multirow{5}{*}{AC2D}
& PhysFTM       & 0.053250 & \textbf{0.088078} & 0.015185 & \textbf{0.089395}\\
& w/o physics   & 0.064986 & \underline{0.091135} & \underline{0.010852} & \underline{0.092520}\\
& w/o RKHS reg. & \underline{0.048491} & 0.096622 & 0.027551 & 0.097985\\
& solve-only        & 1.155543 & 0.171035 & \textbf{0.008657} & 0.173653\\
& gradient-only        & \textbf{$1.76{\times}10^{-5}$} & 0.352111 & 0.351441 & 0.352132\\
\midrule
\multirow{5}{*}{NS2D}
& PhysFTM       & \underline{0.662575} & \textbf{0.060270} & \underline{0.004189} & \textbf{0.061196}\\
& w/o physics   & 1.045838 & \underline{0.085478} & 0.009674 & \underline{0.086781}\\
& w/o RKHS reg. & 1.703260 & 0.113713 & 0.016850 & 0.115421\\
& solve-only        & 1.625978 & 0.105585 & \textbf{0.000653} & 0.107207\\
& gradient-only        & \textbf{0.450487} & 0.090639 & 0.073760 & 0.091112\\
\bottomrule
\end{tabular}
\end{table}

The complete model gives the lowest Full and Off-sensor RMSE on both datasets. Although solve-only achieves the lowest Sensor RMSE, its substantially higher Off-sensor RMSE indicates that accurate fitting at sensor locations does not necessarily translate into accurate reconstruction elsewhere. The gradient-only variant obtains a low PDE residual on AC2D together with a high reconstruction RMSE. These comparisons assess the combined choices of block updates and their effective objectives. Overall, PhysFTM offers the most favorable trade-off among all evaluated variants between sensor fitting, off-sensor reconstruction, and physical consistency. The block updates and active objective terms are specified in Appendix~\ref{app:implementation-details}. Additional spatial and collocation ablations are in Appendix~\ref{app:spatialmodel}.

\noindent\textbf{Reconstruction at non-grid-aligned times.}
We evaluate the learned core trajectory at 100 non-grid-aligned timestamps within the interval spanned by the timestamps of support frames. These evaluation timestamps do not coincide with the original frame timestamps and are therefore distinct from missing frames. For both AC2D and NS2D, the ground truth is recorded directly by rerunning the PDE solver at the perturbed physical timestamps rather than interpolating stored frames. The timestamp and split rules are given in Appendix~\ref{app:continuous-query-protocol}.

\begin{table}[!tbp]

  \centering
  \small
  \setlength{\tabcolsep}{4pt}
  \caption{RMSE at non-grid-aligned times within the continuous support-frame time range, using ground truth directly recorded by the PDE solver. ``Learned'' is the gated ODE-RNN correction to cubic Hermite interpolation in Tucker-core space.}
  \label{tab:continuous-query}
  \begin{tabular}{@{}lcrrr@{}}
    \toprule
    Dataset & $s$ & Learned & Linear & Cubic \\
    \midrule
    AC2D & 0.01 & \textbf{0.180029} & 0.189510 & 0.189556 \\
    AC2D & 0.03 & \textbf{0.085552} & 0.090391 & 0.090388 \\
    NS2D & 0.01 & \textbf{0.130469} & 0.137667 & 0.137453 \\
    NS2D & 0.03 & \textbf{0.055546} & 0.064605 & 0.063790 \\
    \bottomrule
  \end{tabular}
\end{table}

The learned trajectory reduces reconstruction RMSE at non-grid-aligned times relative to both interpolation baselines in all four settings. This experiment supports continuous temporal reconstruction within the interval spanned by the timestamps of support frames on AC2D and NS2D.
\endgroup

%% file: sections/05_conclusion.tex
\section{Conclusion}

\begingroup
We introduce PhysFTM, a functional Tucker method for reconstructing continuous spatiotemporal dynamics from sparse sensors and missing frames. By parameterizing Tucker factors by RKHS and enforcing governing equations via finite-difference residuals, the proposed framework is able to improve the physical consistency and reconstruction accuracy. Experiments on Allen--Cahn, Navier--Stokes, Heat equations, and a gridded Copernicus ocean reanalysis demonstrate accurate and physically consistent reconstruction across 2D and 3D grids.

The main limitations are as follows. First, the current framework considers sparse observations collected from fixed spatial sensors, with missing temporal frames. Extending it to more general observation settings, where measurements are available at arbitrary space-time locations, remains an interesting direction. Second, PhysFTM relies on known governing equations to regularize the reconstruction, which may limit its applicability when the underlying physical models are inaccurate or incomplete. Third, sparse observations generally do not uniquely determine the underlying physical field. Current PhysFTM returns a single regularized field without quantifying the associated reconstruction uncertainty. Therefore, we may extend the method to observations at arbitrary space-time coordinates and incorporate uncertainty quantification in the future to characterize reconstruction ambiguity and provide reliable estimates for downstream analysis.
\endgroup

%% file: sections/06_appendix.tex
\begingroup
\section{Comparison methods and spatial analysis}
\label{app:spatialmodel}

\paragraph{Compared methods.}
SDIFT~\citep{chen2025sdift} is a cross-trajectory diffusion model conditioned on sparse spatiotemporal observations. LRTFR~\citep{luo2024lowrank} fits a functional tensor representation to the sensor coordinate--value measurements of each target trajectory. RecFNO~\citep{zhao2024recfno} is trained across trajectories and maps gridded sparse-field/mask inputs to prescribed-grid outputs. SIREN-MLP~\citep{sitzmann2020siren} and PINNs~\citep{raissi2019physics} are coordinate-to-field controls trained once on the training split and evaluated directly at test coordinates without test-time adaptation; they share the same dense prediction across spatial ratios and temporal protocols, while only the evaluated subsets change. The earlier response used ``SIREN-INR'' for the same SIREN-MLP implementation.

\begin{table*}[t]
  \centering
  \scriptsize
  \setlength{\tabcolsep}{2pt}
  \renewcommand{\arraystretch}{1.08}
  \caption{Task-level comparison protocols. ``Native'' denotes prescribed dataset timestamps; ``coordinate'' denotes direct coordinate evaluation. Baseline budgets and checkpoint selection follow their executed run configurations.}
  \label{app:comparison-protocols}
  \begin{tabular}{@{}p{.07\textwidth}p{.14\textwidth}p{.14\textwidth}p{.15\textwidth}p{.10\textwidth}p{.13\textwidth}p{.15\textwidth}@{}}
    \toprule
    Method & Test input & Training / fitting & Supervision / equation use & Test optimization & Spatial scope & Time interface / selection \\
    \midrule
    SDIFT & Sparse spatiotemporal observations & Cross-trajectory diffusion training & Dense training trajectories; no test missing-frame targets & Posterior sampling & 2D/3D & Native-grid protocols; executed checkpoint \\
    LRTFR & Sparse coordinate--value tuples & Per-trajectory representation fitting & Sensor measurements; no equation loss & Per-trajectory fit & 2D/3D & Native-grid evaluation; fitted objective \\
    SIREN-MLP & Coordinates; no test-mask conditioning & Train once on training split & Training fields; no equation loss & None & 2D/3D & Coordinate output; executed checkpoint \\
    PINNs & Coordinates; no test-mask conditioning & Train once on training split & Training fields and implemented equation residual & None & 2D/3D & Coordinate output; executed checkpoint \\
    RecFNO & Gridded sparse field and mask & Cross-trajectory operator training & Dense training fields; no equation loss & None & 2D/3D & Prescribed-grid support/missing frames; executed checkpoint \\
    PhysFTM & Sparse support-frame sensor values & Cross-trajectory decoder/head training & Observations, missing-frame targets, and equation loss where enabled & Support-core CG solve & 2D/3D & Native-grid and AC/NS in-range non-grid-aligned reconstruction; spatial: training dense RMSE; temporal: validation missing-frame relative $\ell_2$ \\
    \bottomrule
  \end{tabular}
\end{table*}

PhysFTM trains the spatial stage for 500 epochs, where one epoch traverses the complete training loader, and selects the spatial checkpoint by the lowest dense-field training RMSE after inverse normalization, averaging evaluation mini-batch RMSEs. Its temporal head is trained for 20 epochs with AdamW and selects the checkpoint with the lowest validation relative $\ell_2$ error over missing frames, using decoded fields at all spatial locations after inverse normalization. RecFNO is evaluated on prescribed-grid support frames and missing frames.

\begin{table*}[t]
  \centering
  \scriptsize
  \setlength{\tabcolsep}{2.4pt}
  \caption{Additional spatial analysis at $s=0.03$. Panel A evaluates reconstruction at off-sensor locations in support frames. Panel B varies the equation-collocation placement; the complete model uses off-sensor collocation. (\textbf{Bold}: best, \underline{Underline}: second best)}
  \label{tab:appendix-spatial-analysis}

  \begin{tabular}{@{}lrrrrrr@{}}
    \toprule
    \multicolumn{7}{c}{\textbf{Panel A: off-sensor reconstruction in support frames}} \\
    \midrule
    \multirow{2}{*}{Method} & \multicolumn{3}{c}{AC2D} & \multicolumn{3}{c}{NS2D} \\
    \cmidrule(lr){2-4}\cmidrule(lr){5-7}
    & RMSE & SSIM & PDE & RMSE & SSIM & PDE \\
    \midrule
    Linear interp. & 0.1275 & 0.6379 & 0.5877 & 0.0897 & 0.9322 & 1.1605 \\
    Cubic interp.  & 0.1051 & 0.7632 & 0.7174 & \underline{0.0653} & \textbf{0.9656} & 0.8339 \\
    SIREN-MLP      & 0.2907 & 0.1747 & 1.1703 & 0.2155 & 0.5836 & 2.1245 \\
    MLP-FTM        & 0.1049 & 0.7515 & 0.2282 & 0.0859 & 0.9116 & 4.0426 \\
    MLP-FTM+CG     & 0.1026 & 0.7719 & 0.1995 & 0.0832 & 0.9276 & 1.2277 \\
    FTM-RKHS       & \underline{0.0925} & \underline{0.7893} & \underline{0.1532} & 0.0868 & 0.9468 & \underline{0.7371} \\
    PhysFTM        & \textbf{0.0894} & \textbf{0.7938} & \textbf{0.1325} & \textbf{0.0612} & \underline{0.9636} & \textbf{0.3500} \\
    \bottomrule
  \end{tabular}

  \par\vspace{0.8em}

  \begin{tabular}{@{}lrrrr@{}}
    \toprule
    \multicolumn{5}{c}{\textbf{Panel B: equation-collocation placement}} \\
    \midrule
    \multirow{2}{*}{Placement} & \multicolumn{2}{c}{AC2D} & \multicolumn{2}{c}{NS2D} \\
    \cmidrule(lr){2-3}\cmidrule(lr){4-5}
    & RMSE & PDE & RMSE & PDE \\
    \midrule
    w/o equation         & 0.0925 & 0.1532 & 0.0868 & 0.7371 \\
    sensor locations     & 0.0898 & 0.1358 & 0.0622 & 0.3584 \\
    off-sensor locations & \textbf{0.0884} & \textbf{0.1316} & \textbf{0.0609} & \textbf{0.3338} \\
    full domain          & \underline{0.0897} & \underline{0.1344} & \underline{0.0618} & \underline{0.3556} \\
    \bottomrule
  \end{tabular}
\end{table*}

The Panel A comparisons isolate basis parameterization and core optimization, whereas Panel B isolates collocation placement. These diagnostics are separate from the component/optimization ablation in Table~\ref{tab:field-collocation-cvprlike}, whose PDE column uses a disjoint held-out off-sensor set and internal centered-stencil times.

\section{Finite-difference residuals and physical diagnostics}
\label{app:pdeloss}

All derivatives are evaluated after inverse normalization. AC2D uses the Allen--Cahn residual
\[
\mathcal R_{\mathrm{AC}}=D_t\hat u-\lambda_{\mathrm{AC}}\bigl(\epsilon^2(D_{xx}+D_{yy})\hat u-\hat u^3+\hat u\bigr).
\]
NS2D stores $(u,v,\omega)$ rather than pressure. Its held-out residual reuses the centered finite-difference vorticity-equation component of the training operator,
\[
\mathcal R_{\mathrm{NS}}=D_t\hat\omega+\hat uD_x\hat\omega+\hat vD_y\hat\omega-\nu(D_{xx}+D_{yy})\hat\omega-f,
\]
and excludes the first and last temporal points. Training collocation and held-out off-sensor evaluation points are disjoint. Heat3D uses $\mathcal R_{\mathrm{heat}}=D_t\hat u-0.01(D_{xx}+D_{yy}+D_{zz})\hat u$ in the spatial stage with weight $10^{-4}$, while its temporal equation weight is zero. Ocean3D has no governing-equation loss in the reported configuration.

For NS training, the equation-based loss is the sum of three mean squared residuals,
\begin{equation}
\begin{aligned}
\mathcal L_{\mathrm{PDE}}^{\mathrm{NS}}
&=\operatorname{mean}(\mathcal R_{\mathrm{NS}}^2)
+\operatorname{mean}\big[(D_x\hat u+D_y\hat v)^2\big]\\
&\quad+\operatorname{mean}\big[(\hat\omega-(D_x\hat v-D_y\hat u))^2\big],
\end{aligned}
\end{equation}
multiplied by the overall PDE-loss weight. The reported NS equation-residual metric is the mean squared vorticity-equation residual, $\operatorname{mean}(\mathcal R_{\mathrm{NS}}^2)$. The main prediction and ground-truth PDE exports use this same single-equation convention, while the unweighted sum of the three constraints is retained as a separate auxiliary export. The ground-truth export evaluates the numerical equation residual of the reference field.

\subsection{Missing-frame physical diagnostics}
\label{app:physical-diagnostics}

The NS2D diagnostics are evaluated on inverse-normalized full grids of missing frames. Spatial derivatives use centered differences, and Fourier wavenumbers use the physical domain lengths. Divergence measures the predicted velocity field itself, $|\nabla\cdot\hat{\mathbf u}|$. Vorticity is derived independently from predicted and reference velocities and then compared:
\begin{align*}
D_{\mathrm{div}}&=\left[\frac{1}{M}\sum_{n,i,\mathbf x}|\partial_x\hat u+\partial_y\hat v|^2\right]^{1/2},\\
D_{\mathrm{vort}}&=\left[\frac{1}{M}\sum_{n,i,\mathbf x}|(\partial_x\hat v-\partial_y\hat u)-(\partial_xv-\partial_yu)|^2\right]^{1/2}.
\end{align*}
Here the sums pool all test trajectories, their missing frames, and all full-grid points before taking the square root. For the spectrum diagnostic, the spatial means of $(u,v)$ are removed, a radial shell-sum kinetic-energy spectrum $E_{n,i}(k)$ is formed, and the frame error is
\[
e_{n,i}^{\mathrm{spec}}=\frac{\|\widehat E_{n,i}-E_{n,i}\|_2}{\|E_{n,i}\|_2}.
\]
Frame errors are averaged within each trajectory, after which the evaluator computes the mean and standard deviation of the trajectory means. The tables report the mean component.

\begin{table}[t]
  \centering
  \footnotesize
  \setlength{\tabcolsep}{3pt}
  \caption{Aggregation rules for missing-frame physical diagnostics.}
  \label{tab:diagnostic-reducers}
  \begin{tabular}{@{}p{.23\linewidth}p{.46\linewidth}p{.23\linewidth}@{}}
    \toprule
    Diagnostic & Missing-frame/spatial reduction & Trajectory reduction \\
    \midrule
    Divergence error & Global RMS over all missing frames and grid points & Included in the same global pool \\
    Vorticity RMSE & Global RMSE over all missing frames and grid points & Included in the same global pool \\
    Spectrum error & Per-frame relative $L_2$, then missing-frame mean & Mean/std of trajectory means \\
    \bottomrule
  \end{tabular}
\end{table}

\begin{table*}[t]
\centering
\scriptsize
\setlength{\tabcolsep}{2.4pt}
\caption{Complete NS2D missing-frame diagnostics across spatial sensor ratios. Lower is better.(\textbf{Bold}: best, \underline{Underline}: second best)}
\label{tab:full-physical-diagnostics}
\begin{tabular}{@{}cllrrrr@{}}
\toprule
Protocol & $s$ & Method & Off-miss & Divergence & Vort. RMSE & Spectrum \\
\midrule
\multirow{18}{*}{Random}
& 0.01 & SDIFT     & 0.509261 & 0.307569 & 0.876079 & 0.865337 \\
&      & LRTFR     & \underline{0.170679} & 0.475408 & \underline{0.237857} & \textbf{0.048102} \\
&      & PhysFTM   & \textbf{0.130659} & \underline{0.274083} & \textbf{0.232358} & \underline{0.104305} \\
&      & RecFNO    & 0.413223 & 0.514580 & 0.396548 & 0.161395 \\
&      & SIREN-MLP & 0.244360 & 1.886754 & 0.623300 & 0.409266 \\
&      & PINNs     & 0.341072 & \textbf{0.120243} & 0.586433 & 0.402015 \\
& 0.03 & SDIFT     & 0.510274 & 0.352407 & 0.916343 & 0.877309 \\
&      & LRTFR     & 0.145757 & 0.459044 & 0.237923 & \underline{0.049954} \\
&      & PhysFTM   & \textbf{0.064976} & \underline{0.134170} & \textbf{0.109304} & \textbf{0.033599} \\
&      & RecFNO    & \underline{0.104413} & 0.459014 & \underline{0.118608} & 0.067430 \\
&      & SIREN-MLP & 0.244405 & 1.887604 & 0.623477 & 0.408029 \\
&      & PINNs     & 0.340554 & \textbf{0.120243} & 0.586433 & 0.398297 \\
& 0.15 & SDIFT     & 0.510589 & 0.782450 & 1.288744 & 0.882962 \\
&      & LRTFR     & 0.138756 & 0.467318 & \underline{0.236704} & 0.050423 \\
&      & PhysFTM   & \textbf{0.033357} & \textbf{0.029145} & \textbf{0.042455} & \textbf{0.029433} \\
&      & RecFNO    & \underline{0.040051} & 0.460688 & 0.729675 & \underline{0.041657} \\
&      & SIREN-MLP & 0.244476 & 1.887873 & 0.623739 & 0.407336 \\
&      & PINNs     & 0.340193 & \underline{0.120243} & 0.586433 & 0.399324 \\
\midrule
\multirow{18}{*}{Interval}
& 0.01 & SDIFT     & 0.492242 & 0.304216 & 0.881166 & 0.870480 \\
&      & LRTFR     & 0.206649 & 0.889423 & \underline{0.311479} & \textbf{0.069965} \\
&      & PhysFTM   & \textbf{0.164821} & \underline{0.270320} & \textbf{0.251761} & \underline{0.128132} \\
&      & RecFNO    & 0.478397 & 0.374108 & 0.347071 & 0.178211 \\
&      & SIREN-MLP & \underline{0.206503} & 1.858348 & 0.528179 & 0.356109 \\
&      & PINNs     & 0.306602 & \textbf{0.120243} & 0.586433 & 0.346441 \\
& 0.03 & SDIFT     & 0.502748 & 0.344546 & 0.931635 & 0.892291 \\
&      & LRTFR     & \underline{0.196613} & 0.616958 & 0.272195 & \underline{0.055034} \\
&      & PhysFTM   & \textbf{0.150558} & \underline{0.129873} & \textbf{0.150436} & \textbf{0.036346} \\
&      & RecFNO    & 0.359434 & 0.342736 & \underline{0.211528} & 0.055423 \\
&      & SIREN-MLP & 0.213537 & 1.863564 & 0.546894 & 0.368662 \\
&      & PINNs     & 0.318479 & \textbf{0.120243} & 0.586433 & 0.349142 \\
& 0.15 & SDIFT     & 0.485171 & 0.816494 & 1.383403 & 0.925256 \\
&      & LRTFR     & \underline{0.172046} & 0.550371 & 0.254336 & 0.061462 \\
&      & PhysFTM   & \textbf{0.134567} & \textbf{0.026805} & \textbf{0.105232} & \textbf{0.010536} \\
&      & RecFNO    & 0.318191 & 0.321772 & \underline{0.172333} & \underline{0.061395} \\
&      & SIREN-MLP & 0.203190 & 1.853913 & 0.520432 & 0.374622 \\
&      & PINNs     & 0.300163 & \underline{0.120243} & 0.586433 & 0.373544 \\
\bottomrule
\end{tabular}
\end{table*}

\endgroup

\section{Theoretical analysis on spatial reconstruction}\label{app:Theory}
\begingroup
In this section, we conduct an idealized spatial reconstruction error analysis for the RKHS-parameterized Functional Tucker Model (FTM-RKHS) under sparse observations. We aim to characterize how the modewise truncation level, kernel spectrum, and number of measurement samples jointly determine the reconstruction error. The i.i.d. spatial-sampling assumption is an analytical idealization and is not identical to the fixed-within-trajectory sensor protocol used in the experiments. Moreover, the spectral truncation level $r$ in this analysis is not identified with the implemented Tucker rank.
\endgroup

Let \(u:\mathcal Z\to\mathbb R\) denote the target field, where \(\mathcal Z\) contains trajectory, time, spatial coordinates, and channel indices. The field is measured only on a sparse set \(\Omega=\{z_i\}_{i=1}^n\), with \(n=|\Omega|\), through measurements \(y_i=u(z_i)+\varepsilon_i\), where \(\mathbb E[\varepsilon_i\mid z_i]=0\). The task is to recover the field $u$ via the measurements $y_i$.

For the sake of theoretical analysis, we consider the following assumptions. For high-probability bounds, we assume that the noise is conditionally sub-Gaussian, i.e., $\mathbb E\left[\exp(t\varepsilon_i)\mid z_i\right]\leq\exp\left(\frac{\sigma^2t^2}{2}\right),\ t\in\mathbb R.$
We assume that the observation distribution echoes the target distribution \(\rho\) over the full space \(\mathcal Z\), i.e., the observations $z_i$ are sampled i.i.d. from $\rho$.

\paragraph{From one-dimensional RKHS to tensor-product RKHS.}

We first recall the one-dimensional RKHS construction for each mode, following the standard spectral characterization of RKHSs induced by compact positive kernel integral operators \citep{aronszajn1950theory,berlinet2004reproducing,steinwart2008support}. Let \((\mathcal Z_j,\mathcal B_j,\rho_j)\), \(j=1,\dots,d\), be the probability space associated with the \(j\)-th mode, where \(\mathcal Z_j\) is the input domain, \(\mathcal B_j\) is the corresponding sigma-algebra, and \(\rho_j\) is the probability measure on \(\mathcal Z_j\). Let \(\mathcal H_j\) be an RKHS on \(\mathcal Z_j\) with kernel \(k_j\), the kernel integral operator is defined by \((T_j f)(z_j)=\int_{\mathcal Z_j}k_j(z_j,z_j')f(z_j')\,d\rho_j(z_j')\).
Assume that each \(T_j: L^2(\rho_j)\to L^2(\rho_j)\) is compact, self-adjoint, and positive semidefinite, so that the Mercer-type spectral decomposition applies \citep{steinwart2008support,cucker2007learning}. Then, there exist eigenpairs \(T_j\phi_{j,m}=\lambda_{j,m}\phi_{j,m}\), with \(\lambda_{j,1}\geq \lambda_{j,2}\geq\cdots>0\), where \(\{\phi_{j,m}\}_{m\geq1}\) is orthonormal in \(L^2(\rho_j)\).

For \(f_j=\sum_{m\geq1}a_{j,m}\phi_{j,m}\) and \(g_j=\sum_{m\geq1}b_{j,m}\phi_{j,m}\), the one-dimensional \(L^2\) and RKHS inner products are defined by
\[
\langle f_j,g_j\rangle_{L^2(\rho_j)}
:=
\sum_{m\geq1}a_{j,m}b_{j,m},
\qquad
\langle f_j,g_j\rangle_{\mathcal H_j}
:=
\sum_{m\geq1}
\frac{a_{j,m}b_{j,m}}{\lambda_{j,m}},
\]
which induces the norms \(\|f_j\|_{L^2(\rho_j)}^2=\sum_{m\geq1}a_{j,m}^2\) and \(\|f_j\|_{\mathcal H_j}^2=\sum_{m\geq1}a_{j,m}^2/\lambda_{j,m}\). Equivalently, \(\{\sqrt{\lambda_{j,m}}\phi_{j,m}\}_{m\geq1}\) forms an orthonormal system in \(\mathcal H_j\).

We now extend the construction to the tensor setting using the standard tensor-product RKHS construction \citep{aronszajn1950theory,berlinet2004reproducing}. Let \(\mathcal Z=\mathcal Z_1\times\cdots\times\mathcal Z_d\), and assume that the full coordinate distribution is \(\rho=\rho_1\otimes\cdots\otimes\rho_d\). For the sake of brevity, we assume that the observation distribution echoes the target distribution.

Define the tensor-product RKHS \(\mathcal H=\bigotimes_{j=1}^d\mathcal H_j\) with tensor-product kernel \(K(z,z')=\prod_{j=1}^d k_j(z_j,z_j')\), which is positive definite by the closure properties of reproducing kernels under tensor products \citep{aronszajn1950theory,berlinet2004reproducing}. For a multi-index \(\boldsymbol m=(m_1,\dots,m_d)\in\mathbb N^d\), define
\[
\Phi_{\boldsymbol m}(z)
=
\prod_{j=1}^d\phi_{j,m_j}(z_j),
\qquad
\Lambda_{\boldsymbol m}
=
\prod_{j=1}^d\lambda_{j,m_j}.
\]
Because \(\rho\) is a product measure, it holds that \(\langle \Phi_{\boldsymbol m},\Phi_{\boldsymbol n}\rangle_{L^2(\rho)}=\prod_{j=1}^d\delta_{m_jn_j}=\delta_{\boldsymbol m\boldsymbol n}\), i.e., \(\{\Phi_{\boldsymbol m}\}_{\boldsymbol m\in\mathbb N^d}\) is orthonormal in \(L^2(\rho)\).

The tensor-product integral operator is \(T=\bigotimes_{j=1}^dT_j\), and the preceding definitions imply \(T\Phi_{\boldsymbol m}=\Lambda_{\boldsymbol m}\Phi_{\boldsymbol m}\). Thus, the tensor spectrum follows from the one-dimensional spectra and the product-measure structure.

For two functions \(f=\sum_{\boldsymbol m}a_{\boldsymbol m}\Phi_{\boldsymbol m}\) and \(g=\sum_{\boldsymbol m}b_{\boldsymbol m}\Phi_{\boldsymbol m}\), the tensor-product RKHS inner product is given by
\[
\langle f,g\rangle_{\mathcal H}
=
\sum_{\boldsymbol m\in\mathbb N^d}
\frac{a_{\boldsymbol m}b_{\boldsymbol m}}{\Lambda_{\boldsymbol m}}
=
\sum_{\boldsymbol m\in\mathbb N^d}
\frac{a_{\boldsymbol m}b_{\boldsymbol m}}
{\prod_{j=1}^d\lambda_{j,m_j}},
\]
which introduces the norm
\[
\|f\|_{L^2(\rho)}^2
=
\sum_{\boldsymbol m\in\mathbb N^d}a_{\boldsymbol m}^2,
\qquad
\|f\|_{\mathcal H}^2
=
\sum_{\boldsymbol m\in\mathbb N^d}
\frac{a_{\boldsymbol m}^2}{\Lambda_{\boldsymbol m}}.
\]

\paragraph{Truncated RKHS-Tucker function class.}

For a fixed truncation level \(r\), define the rectangular modewise truncation region
\[
\mathcal R_r
=
\left\{
\boldsymbol m\in\mathbb N^d:
1\leq m_j\leq r,\ \forall j=1,\dots,d
\right\}.
\]
Its complement is \(\mathbb N^d\setminus\mathcal R_r=\{\boldsymbol m\in\mathbb N^d:\exists j,\ m_j>r\}\), which contains all index tuples where at least one mode enters the high-frequency region $\{r+1,r+2,\dots\}$.

Using the tensor-product bases defined above, the truncated RKHS-Tucker function class is
\[
\begin{aligned}
\mathcal F_r=\biggl\{f:\mathcal Z\to\mathbb R\ \biggm|\
&f(z)=\sum_{\boldsymbol m\in\mathcal R_r}\theta_{\boldsymbol m}\Phi_{\boldsymbol m}(z),\\
&\Theta=(\theta_{\boldsymbol m})_{\boldsymbol m\in\mathcal R_r}
\text{ admits a Tucker low-rank structure}\biggr\}.
\end{aligned}
\]

Let \(u_r^\star\) be the best approximation of \(u\) within \(\mathcal F_r\), namely
\[
u_r^\star
=
\arg\min_{f\in\mathcal F_r}
\mathcal R(f),
\qquad
\mathcal R(f)
=
\mathbb E_{z\sim\rho}
\left[
(f(z)-u(z))^2
\right].
\]
The empirical estimator is denoted by \(\hat u_r\). The total reconstruction error satisfies
\[
\|\hat u_r-u\|_{L^2(\rho)}
\leq
\underbrace{\|u_r^\star-u\|_{L^2(\rho)}}_{\mathrm{approximation\ error}}
+
\underbrace{\|\hat u_r-u_r^\star\|_{L^2(\rho)}}_{\mathrm{statistical\ error}}.
\]

\paragraph{Approximation error.}
We evaluate the approximation error by using a specific RKHS-Tucker function in ${\cal F}_r$. Assume that the target function satisfies the following tensor source condition, which is a standard smoothness assumption in spectral regularization and kernel learning theory \citep{caponnetto2007optimal,cucker2007learning,steinwart2009optimal}:
\[
u=T^v g,
\qquad
\|g\|_{L^2(\rho)}\leq C_g,
\qquad
v>\frac12.
\]
Using the spectral notation defined above, it holds that \(g=\sum_{\boldsymbol m}g_{\boldsymbol m}\Phi_{\boldsymbol m}\) and \(u=\sum_{\boldsymbol m}\Lambda_{\boldsymbol m}^v g_{\boldsymbol m}\Phi_{\boldsymbol m}\). Let \(P_r^{\otimes d}\) denote the tensor-product projection onto the span of bases indexed by \(\mathcal R_r\). By the RKHS norm, the truncation error can be given by
\[
\mathcal E_{\mathrm{app},\mathcal H}^2(r)
:=
\|(I-P_r^{\otimes d})u\|_{\mathcal H}^2
=
\sum_{\boldsymbol m\notin\mathcal R_r}
\Lambda_{\boldsymbol m}^{2v-1}
g_{\boldsymbol m}^2.
\]
The exponent \((2v-1)\) appears because the RKHS norm divides the squared coefficient by the corresponding kernel eigenvalue.

Assume further that each mode has polynomial eigenvalue decay \(\lambda_{j,m}\leq K_0m^{-\beta}\), a standard capacity condition for kernel integral operators \citep{caponnetto2007optimal,steinwart2009optimal}. Let \(\gamma=\beta(2v-1)\). If \(\gamma>1\), then the one-dimensional spectral tail satisfies \(\sum_{m>r}\lambda_{j,m}^{2v-1}\lesssim r^{1-\gamma}\), while the corresponding low-frequency sum is uniformly bounded in \(r\). Therefore, the truncation error satisfies
\[
\mathcal E_{\mathrm{app},\mathcal H}^2(r)
\lesssim
C r^{1-\beta(2v-1)},
\]
or equivalently,
\[
\mathcal E_{\mathrm{app},\mathcal H}(r)
\lesssim
C_A r^{-s},
\qquad
s=
\frac{\beta(2v-1)-1}{2}.
\]
Here \(C_A\) absorbs the source radius \(C_g\), the spectral decay constant \(K_0\), the tensor order \(d\), and the bounded low-frequency spectral sums; in particular, \(C_A\) is independent of \(r\). Under the continuous embedding \(\|h\|_{L^2(\rho)}\leq C_{\mathrm{emb}}\|h\|_{\mathcal H}\), the same rate gives
\[
\|u_r^\star-u\|_{L^2(\rho)}\leq C_{\mathrm{emb}}\mathcal E_{\mathrm{app},{\cal H}}(r)
\lesssim
C_A r^{-s},
\qquad
s=
\frac{\beta(2v-1)-1}{2}.
\]
where \(C_A\) additionally absorbs \(C_{\mathrm{emb}}\).

\paragraph{Statistical error.}

For the statistical component, define the truncated tensor effective dimension as a finite-dimensional analogue of the effective dimension used in kernel ridge regression and regularized least squares \citep{caponnetto2007optimal,steinwart2009optimal}:
\[
d_{\mathrm{eff},r}^{\mathrm{tensor}}(\lambda)
=
\sum_{m_1\leq r,\dots,m_d\leq r}
\frac{
\prod_{j=1}^d\lambda_{j,m_j}
}{
\prod_{j=1}^d\lambda_{j,m_j}+\lambda
}.
\]
Here \(\lambda>0\) is the regularization parameter. It determines the learnability of each tensor spectral direction: directions with \(\prod_{j=1}^d\lambda_{j,m_j}\gg\lambda\) are effectively retained, while directions with \(\prod_{j=1}^d\lambda_{j,m_j}\ll\lambda\) are suppressed by regularization. Thus, \(d_{\mathrm{eff},r}^{\mathrm{tensor}}(\lambda)\) measures the effective number of learnable tensor spectral directions within the truncated RKHS class. Equivalently, the same statistical term can be obtained through localized Rademacher complexity for squared-loss regression under boundedness and local curvature conditions \citep{bartlett2002rademacher,bartlett2005local}.

Under bounded-kernel, sub-Gaussian noise, and local curvature assumptions, standard kernel regression and localized Rademacher arguments yield that, with probability at least \(1-\delta\) \citep{caponnetto2007optimal,bartlett2005local},
\[
\mathcal E_{\mathrm{stat}}(r,n,\delta)
=
\|\hat u_r-u_r^\star\|_{L^2(\rho)}
\lesssim
C_S
\sqrt{
\frac{
d_{\mathrm{eff},r}^{\mathrm{tensor}}(\lambda)
+
\log(1/\delta)
}{n}
}.
\]
Since \(d_{\mathrm{eff},r}^{\mathrm{tensor}}(\lambda)\leq r^d\), a conservative rank-based bound is
\[
\mathcal E_{\mathrm{stat}}(r,n,\delta)
\lesssim
C_S
\sqrt{
\frac{
r^d+\log(1/\delta)
}{n}
}.
\]

\paragraph{Balancing reconstruction rate.}
Combining the approximation and statistical bounds, we yield
\[
\|\hat u_r-u\|_{L^2(\rho)}
\lesssim
C_A r^{-s}
+
C_S
\sqrt{
\frac{
r^d+\log(1/\delta)
}{n}
},
\qquad
s=
\frac{\beta(2v-1)-1}{2}.
\]
Ignoring the lower-order concentration term, the optimal truncation level is obtained by balancing \(C_A r^{-s}\asymp C_S r^{d/2}n^{-1/2}\), which gives
\[
r_\star
\asymp
n^{\frac{1}{2s+d}}.
\]
Substituting \(r_\star\) into the total error, the right-hand side boils down to $C n^{-\frac{s}{2s+d}}$.

Since \(2s=\beta(2v-1)-1\), we obtain
\[
\|\hat u_{r_\star}-u\|_{L^2(\rho)}
\lesssim
C
n^{
-\frac{\beta(2v-1)-1}
{2\left(\beta(2v-1)-1+d\right)}
}
+
C
\sqrt{
\frac{\log(1/\delta)}{n}
}.
\]
This rate reflects two effects: the kernel spectrum controls how fast the rectangular truncation error decays, while the retained tensor-product spectral dimension controls the statistical term.

\paragraph{Sample complexity.}
Finally, the sample complexity can be inferred from the balanced rate. To ensure that \(\|\hat u_r-u\|_{L^2(\rho)}\leq\epsilon\) with probability at least \(1-\delta\), it suffices to choose
\[
n
\gtrsim \left(\frac{2C}{\epsilon}\right)^{
\frac{
2\left(\beta(2v-1)-1+d\right)
}{
\beta(2v-1)-1
}
}
+
\frac{4C^2\log(1/\delta)}{\epsilon^2}.
\]

\paragraph{Specialization to the Mat\'ern-\(3/2\) kernel.}
We now adopt the preceding bound to the Mat\'ern-\(3/2\) kernel. For a one-dimensional mode, the Mat\'ern kernel with smoothness parameter \(\nu_{\mathrm M}=3/2\) has polynomial spectral decay \(\lambda_{j,m}\lesssim m^{-\beta}\), with \(\beta=2\nu_{\mathrm M}+1=4\), under the standard one-dimensional Sobolev/Mat\'ern spectral correspondence \citep{rasmussen2006gaussian,stein1999interpolation}. The value of \(\beta\) is determined by the kernel spectrum, whereas the source exponent \(v\) describes the regularity of the target field relative to the kernel integral operator.

If the target field satisfies the second-order source condition \(u=T^2g\) with \(\|g\|_{L^2(\rho)}\leq C_g\), then \(v=2\), so \(\gamma=\beta(2v-1)=12\) and \(s=(\gamma-1)/2=11/2\). The approximation and total reconstruction bounds become
\[
\mathcal E_{\mathrm{app}}(r)
\lesssim
C_A r^{-11/2},
\qquad
\|\hat u_r-u\|_{L^2(\rho)}
\lesssim
C_A r^{-11/2}
+
C_S
\sqrt{
\frac{
r^d+\log(1/\delta)
}{n}
}.
\]
Balancing \(r^{-11/2}\asymp r^{d/2}n^{-1/2}\) gives \(r_\star\asymp n^{1/(11+d)}\). Substituting this choice yields
\[
\|\hat u_{r_\star}-u\|_{L^2(\rho)}
\lesssim
C
n^{-\frac{11}{2(11+d)}}
+
C
\sqrt{
\frac{\log(1/\delta)}{n}
}.
\]
This rate shows that, under a Mat\'ern-\(3/2\) kernel and a second-order source condition, the approximation term decays rapidly with the truncation level, while the statistical term still depends on the effective tensor dimension induced by the retained modes.

\begingroup
\section{Dataset Information}
\label{app:dataset}

The experiments involve three synthetic benchmarks and a gridded ocean reanalysis. Table~\ref{tab:dataset-details} summarizes the tensors from the datasets.

\begin{table*}[t]
  \centering
  \footnotesize
  \setlength{\tabcolsep}{4pt}
  \caption{Datasets used in the experiments. Shapes are trajectories $\times$ time $\times$ spatial axes $\times$ channels.}
  \label{tab:dataset-details}
  \begin{tabular}{@{}lclll@{}}
    \toprule
    Dataset & Split & Stored tensor & Channels & Domain and time span \\
    \midrule
    AC2D & 160/20/20 & $200\times201\times256\times256$ & scalar $u$ & $[0,1)^2$, $t\in[0,2]$ \\
    NS2D & 640/80/80 & $800\times201\times128\times128\times3$ & $(u,v,\omega)$ & $[0,1)^2$, $t\in[0,20]$ \\
    Heat3D & 160/20/20 & $200\times50\times64\times64\times64\times1$ & temperature $u$ & $[0,1]^3$, $t\in[0,4]$ \\
    Ocean3D & 80/10/10 & $100\times90\times32\times64\times64\times2$ & $(u,v)$ & depth--latitude--longitude, 90 days \\
    \bottomrule
  \end{tabular}
\end{table*}

\paragraph{AC2D and NS2D.}
AC2D solves the periodic two-dimensional Allen--Cahn equation with $\lambda=1$ and $\epsilon=0.01$ and saves 201 frames at interval $0.01$. NS2D solves the forced periodic vorticity equation with viscosity $10^{-4}$, saves 201 frames at interval $0.1$, and stores $(u,v,\omega)$ after downsampling to $128^2$. The AC2D and NS2D training and held-out equation-residual evaluators use centered differences in both space and time and discard the first and last time points.

\paragraph{Heat3D.}
Heat3D contains 200 independent scalar-temperature trajectories. The trajectories are permuted with seed 123 and split 160/20/20. All trajectories share 50 physical save times in $[0,4]$; these times are slightly nonuniform because the explicit-Euler step and the save interval are rounded, while the logical model coordinate is a uniform 50-point grid in $[0,1]$. All $64^3$ voxels are valid, the spatial grid includes both endpoints, and the generator imposes homogeneous Dirichlet conditions on all six faces.

\paragraph{Ocean3D.}
Ocean3D is constructed from daily-mean horizontal currents \((u,v)\) in the CMEMS product \texttt{GLOBAL\_MULTIYEAR\_PHY\_001\_030}. The local source consists of annual NetCDF files, starting with \texttt{uv\_fulldepth\_1993-01-01\_to\_1993-12-31.nc}. Files are ordered chronologically, and each file is divided into nonoverlapping 90-day sequences beginning at its first record. Trailing records that do not form a complete sequence are discarded; sequences never cross file boundaries. We retain the first 100 sequences, comprising 9000 daily fields from 1993 to 2017 and split them sequentially into 80 training, 10 validation, and 10 test sequences. The first and last retained dates of these splits are 1993-01-01 to 2012-12-25, 2013-01-01 to 2015-06-29, and 2015-06-30 to 2017-12-26, respectively; these endpoint ranges do not imply uninterrupted daily coverage because 5 or 6 trailing days are omitted in each year. We directly retain the source-grid crop of $64\times64$ horizontal points spanning latitudes [-2.6666667461, 2.5833332539] and longitudes [167.3333282471, 172.5833282471] degrees, without horizontal interpolation. We select zero-based depth indices 0 to 31; these 32 nonuniform levels range from 0.494024992 to 541.088928 m.Each sequence has shape $90\times32\times64\times64\times2$.

\subsection{Evaluation subsets}

\begin{figure}[!tbp]

  \centering
  \includegraphics[width=0.98\linewidth]{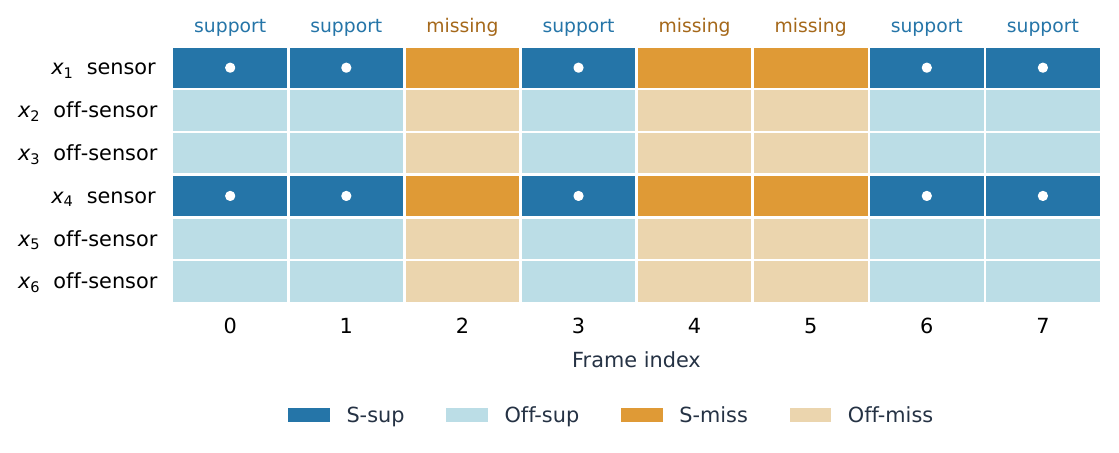}
  \caption{Four evaluation subsets defined by sensor/off-sensor locations and support/missing frames. White dots indicate sensor measurements provided to the model.}
  \label{app:fig:eval_subsets}
  \vspace{-0.8em}
\end{figure}

\label{app:fourway-subsets}

For trajectory $b$, let $\mathcal X_b$ be its evaluation grid, $\mathcal X_{\mathrm{sen}}^{(b)}\subset\mathcal X_b$ its fixed set of sensor locations, and $\mathcal I_{\mathrm{sup}}^{(b,p)}$ and $\mathcal I_{\mathrm{miss}}^{(b,p)}$ the indices of support frames and missing frames under protocol $p$, respectively. The four disjoint subsets are
\begin{align*}
\mathcal A_{\mathrm{S\text{-}sup}}^{(b,p)}&=\mathcal I_{\mathrm{sup}}^{(b,p)}\times\mathcal X_{\mathrm{sen}}^{(b)}, &
\mathcal A_{\mathrm{Off\text{-}sup}}^{(b,p)}&=\mathcal I_{\mathrm{sup}}^{(b,p)}\times(\mathcal X_b\setminus\mathcal X_{\mathrm{sen}}^{(b)}),\\
\mathcal A_{\mathrm{S\text{-}miss}}^{(b,p)}&=\mathcal I_{\mathrm{miss}}^{(b,p)}\times\mathcal X_{\mathrm{sen}}^{(b)}, &
\mathcal A_{\mathrm{Off\text{-}miss}}^{(b,p)}&=\mathcal I_{\mathrm{miss}}^{(b,p)}\times(\mathcal X_b\setminus\mathcal X_{\mathrm{sen}}^{(b)}).
\end{align*}
Support frames provide sparse measurements only at sensor locations, not complete spatial fields. Missing frames provide no sensor measurements to the model; their ground-truth fields, including values at sensor locations, are retained for evaluation. Let \(C_D\) denote the number of channels. For each evaluation subset, we pool squared errors across trajectories, selected space–time points, and channels, divide by the total number of selected scalar values, and take the square root. An example is shown in Figure~\ref{app:fig:eval_subsets}.
\[
\operatorname{RMSE}_{D}(\mathcal A)=\left[\frac{\sum_b\sum_{(t,\mathbf x)\in\mathcal A^{(b)}}\sum_{c=1}^{C_D}|\hat u_b(t,\mathbf x,c)-u_b(t,\mathbf x,c)|^2}{C_D\sum_b|\mathcal A^{(b)}|}\right]^{1/2}.
\]

\subsection{Missing-frame protocols}
\label{app:native-time-protocols}

\begin{figure}[!tbp]

  \centering
  \includegraphics[width=0.98\linewidth]{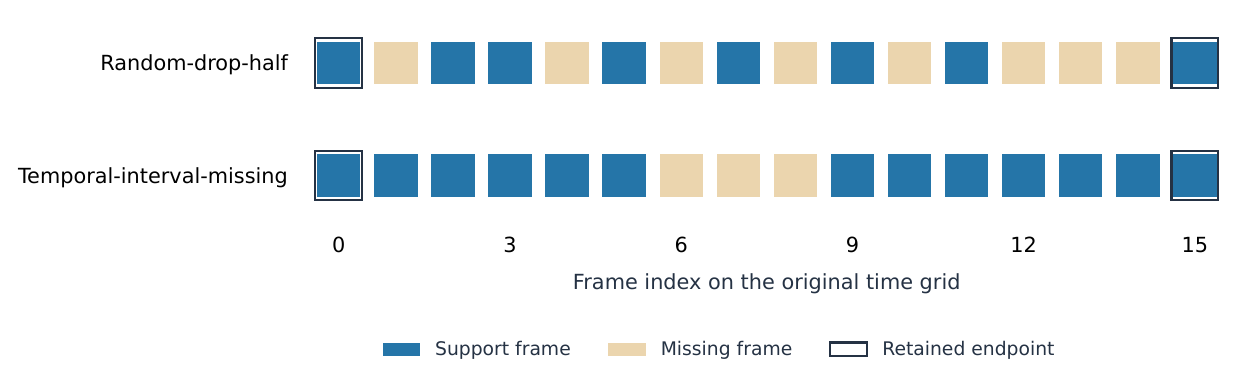}
  \caption{Two missing-frame protocols on the original time grid: randomly withholding frames or withholding a contiguous block. Both endpoints remain support frames.}
  \label{app:fig:missing_protocols}
  \vspace{-0.8em}
\end{figure}

Let $T$ be the max timestamp of the trajectory. Let $M_t^{(b)}$ be the spatial mask fixed within the trajectory, and let $K_t^{(b,p)}\in\{0,1\}$ be a frame gate specific to the trajectory. The observation mask is $\widetilde M_t^{(b,p)}=K_t^{(b,p)}M_t^{(b)}$; when $K_t=1$, the frame is a support frame with sparse sensor measurements, and when $K_t=0$, it is a missing frame with all sensor measurements withheld across depths and channels. This mask affects the input of the model, not the availability of ground truth for evaluation.

\paragraph{Random-drop-half.}
Both endpoints are retained, while internal frames are sampled independently:
\[
K_0^{(b)}=K_{T-1}^{(b)}=1,\qquad K_t^{(b)}\overset{\mathrm{i.i.d.}}{\sim}\operatorname{Bernoulli}(0.5),\quad t=1,\ldots,T-2.
\]
The implementation uses \texttt{numpy.random.RandomState} and split seed $10^5\!\cdot123+10g(s)+i_{\mathrm{split}}$, where $g(0.01)=1$, $g(0.03)=3$, $g(0.15)=15$, and $i_{\mathrm{split}}\in\{0,1,2\}$.

\paragraph{Temporal-interval-missing.}
A contiguous internal block is sampled independently for each trajectory:
\[
L_{\mathrm{miss}}=\max(1,\operatorname{round}(0.2T)),\qquad a_b\sim\operatorname{Uniform}\{1,\ldots,T-L_{\mathrm{miss}}-1\},
\]
and $K_t^{(b)}=0$ for $t\in[a_b,a_b+L_{\mathrm{miss}})$. Both endpoints remain support frames.An example is shown in Figure~\ref{app:fig:missing_protocols}.

\subsection{Reconstruction at non-grid-aligned times}
\label{app:continuous-query-protocol}

\begin{figure}[!tbp]

  \centering
  \includegraphics[width=0.98\linewidth]{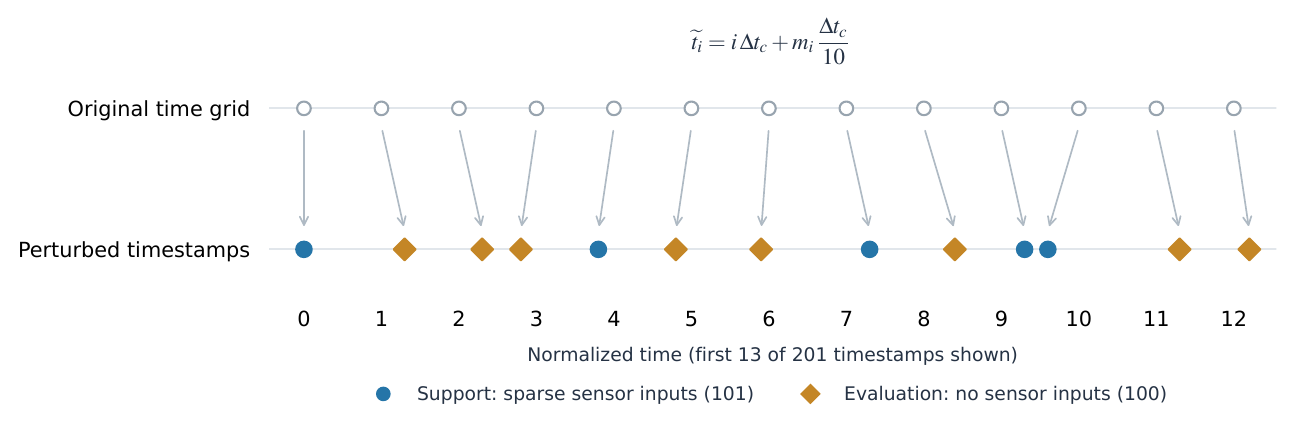}
  \caption{Internal timestamps are perturbed before being split into support and evaluation sets, while both endpoints remain fixed. Evaluation ground truth is recorded directly by the PDE solver.}
  \label{app:fig:non_grid_times}
  \vspace{-0.8em}
\end{figure}

This AC2D/NS2D ablation is separate from the native-grid missing-frame protocols. Its evaluation timestamps are non-grid-aligned relative to the original coarse time grid and lie within the interval spanned by the timestamps of support frames. They are not defined as missing frames from the original stored sequence. Starting from 201 coarse times $t_i^{(0)}=i\Delta t_c$, both endpoints are fixed and every internal time is perturbed by
\[
\widetilde t_i=i\Delta t_c+m_i\frac{\Delta t_c}{10},\qquad
m_i\overset{\mathrm{i.i.d.}}{\sim}\operatorname{Uniform}\{-4,-3,-2,-1,1,2,3,4\},\quad i=1,\ldots,199.
\]
AC2D uses $\Delta t_c=0.01$ on $[0,2]$, and NS2D uses $\Delta t_c=0.1$ on $[0,20]$.100 evaluation indices are sampled without replacement from $\{1,\ldots,199\}$:
\[
I_e\subset\{1,\ldots,199\},\quad |I_e|=\lfloor201\times0.5\rfloor=100,\qquad I_s=\{0,\ldots,200\}\setminus I_e,\quad |I_s|=101.
\]
Sensor measurements are provided only at the timestamps indexed by $I_s$; the timestamps indexed by $I_e$ are used for reconstruction evaluation without providing their sensor measurements to the model. The endpoints therefore always belong to the support set. AC2D and NS2D use the same normalized offset vector and index-level partition between support frames and evaluation snapshots at non-grid-aligned times, shared across all test trajectories and sensors. Ground-truth states are saved directly by rerunning each PDE solver at the perturbed timestamps, not by interpolating uniformly saved snapshots. An example is shown in Figure~\ref{app:fig:non_grid_times}.

\input{tables/appendix_full_reconstruction}

\section{Implementation Details}
\label{app:implementation-details}

The spatial decoder uses a Mat\'ern-$3/2$ kernel. Each active spatial mode has its own factor matrix; two-dimensional data use two modes, while Heat3D and Ocean3D use three. The number of inducing points along each mode equals that mode's grid side length in the reported default configuration; the reported AC2D sensitivity baseline therefore uses 256 inducing points. Spatial factors are shared across channels, with the channel index stored in the Tucker core.

The spatial stage uses batch size 8 and is trained for 500 epochs, where one epoch traverses the complete training loader. For each mini-batch, the conditionally linear cores are updated by conjugate gradients and the RKHS factors are updated with AdamW. The learning rate is \(2\times10^{-4}\); the RKHS, core-ridge, core-TV, and temporal-GP weights are \(10^{-4}\), \(10^{-2}\), \(10^{-2}\), and \(10^{-5}\), respectively. Core updates use five CG iterations and tolerance \(10^{-3}\).

Spatial checkpoints are selected by the lowest dense-field RMSE on the training trajectories after inverse normalization, averaging the RMSE values of evaluation mini-batches. Dense training fields are used for spatial checkpoint selection; spatial reconstruction gradients use only available sensor measurements, together with the stated regularizers and equation constraints.

The complete model samples equation-collocation centers from off-sensor locations on the spatial grid. For each trajectory, the set of sensor locations is the union of spatial locations with sensor measurements at any saved time and in any channel; the eligible collocation set is its spatial complement. AC2D and NS2D use two-dimensional grid centers, while Heat3D uses three-dimensional voxel centers. For Heat3D, centers whose finite-difference stencils would require moving the center inward from a boundary are excluded before sampling. The off-sensor restriction applies to collocation centers; neighboring stencil evaluations may lie at sensor locations. Ocean3D uses no governing-equation loss in the reported setting. AC2D and NS2D use centered spatial and temporal differences and exclude the temporal endpoints from residual training and evaluation.

Before each AdamW factor update, the gradient of each RKHS coefficient matrix is left-multiplied by the inverse Gram matrix of its spatial-mode kernel, \(K_d^{-1}\nabla_{A_d}\mathcal L\). The spatial AdamW optimizer uses weight decay \(10^{-2}\); the gradient-only core updater uses the same weight decay.

We compare block-update strategies with their corresponding effective objectives. The complete model alternates CG core solves with gradient-based RKHS-factor updates. The solve-only variant retains the same ridge-TV core subproblem, but updates each spatial factor by solving for target factor values with a value-ridge penalty and projecting them to RKHS coefficients with kernel ridge regression. These factor updates use observational least squares and kernel projection. The gradient-only variant updates both cores and factors with AdamW using the RMSE over available sensor measurements, RKHS penalty, equation-based loss, and time-kernel penalty. The RKHS penalty depends only on the factors, whereas the time-kernel penalty depends only on the cores. Core-ridge and temporal-TV penalties belong to the CG subproblem. The time-kernel term is defined in Eq.~\eqref{eq:time-gp}.

Fields are normalized with channel-wise train-set statistics. Validation and test splits reuse these statistics, and all reported RMSEs, residuals, and physical diagnostics are computed after inverse normalization. The Ocean3D RecFNO RMSE export originally aggregates both channels jointly; the values reported in Table~\ref{tab:end-to-end-cvprlike} use the corresponding per-channel convention by dividing those four exported RMSE columns by $\sqrt{2}$.

At inference, the spatial decoder is frozen and the core-correction sequence is solved around the train-set mean-core reference on the complete native time grid with ridge $10^{-2}$, TV weight $10^{-2}$, 30 CG iterations, tolerance $10^{-4}$, and batch size 32. Missing frames have zero observation masks; after the joint solve, the cores at the timestamps of support frames are extracted as support anchors.

The residual ODE-RNN temporal head operates in Tucker-core space, uses hidden dimension 256, and is trained for 20 epochs with AdamW (batch size 4, learning rate \(10^{-3}\), and weight decay \(10^{-6}\), except that the learned residual-scale parameter has zero weight decay). Let \(B_i(t)\) be the cubic Hermite baseline constructed from the inferred support cores, and let \(C_\psi(t)\) denote the actual correction added to that baseline, including the interval gate and the learned bounded residual scale. At the timestamps of missing frames, we minimize the mean squared error between \(C_\psi(t)\) and \(G_i^{\mathrm{train}}(t)-B_i(t)\), where \(G_i^{\mathrm{train}}\) is the frozen training-core sequence saved by the spatial stage. Equivalently, the objective is the missing-frame MSE between the predicted core \(B_i(t)+C_\psi(t)\) and \(G_i^{\mathrm{train}}(t)\). The core-loss weight is 1; decoded-field, spatial-gradient, and temporal-equation losses are disabled. We select the checkpoint with the lowest validation relative \(\ell_2\) error over missing frames, computed from decoded fields after inverse normalization.
\begin{equation}
\mathcal L_{\mathrm{temporal}}
=\operatorname{MSE}_{i,t\in\mathcal T_{\mathrm{miss}}^{(i)}}
\left(C_\psi(t),G_i^{\mathrm{train}}(t)-B_i(t)\right).
\end{equation}
The MSE averages all scalar coefficients of the selected cores. Missing-frame validation uses all spatial locations at the selected times.
\endgroup

%% file: tables/appendix_full_reconstruction.tex
\section{Complete reconstruction results across sensor ratios}
\label{app:full-reconstruction-results}

Tables~\ref{tab:full-reconstruction-ac2d}--\ref{tab:full-reconstruction-ocean} report the complete three-ratio four-subset comparison. Rankings, when discussed, use only available values within the same dataset, temporal protocol, spatial ratio, and subset.

\begin{table}[p]
\centering
  \scriptsize
  \setlength{\tabcolsep}{3pt}
  \caption{Complete four-subset reconstruction RMSE on AC2D across spatial sensor ratios. Lower is better. (\textbf{Bold}: best, \underline{Underline}: second best)}
  \label{tab:full-reconstruction-ac2d}
  \begin{tabular}{@{}cllrrrr@{}}
    \toprule
    Protocol & $s$ & Method & S-sup & Off-sup & S-miss & Off-miss \\
    \midrule
    \multirow{18}{*}{Random-drop-half} & \multirow{6}{*}{0.01} & SDIFT & 0.373381 & 0.381700 & 0.378977 & 0.381200 \\
     &  & LRTFR & 0.311380 & \underline{0.329100} & 0.308521 & \underline{0.328814} \\
     &  & PhysFTM & \textbf{0.009898} & \textbf{0.187900} & \underline{0.042055} & \textbf{0.191673} \\
     &  & RecFNO & \underline{0.020782} & 0.697782 & \textbf{0.021170} & 0.696267 \\
     &  & SIREN-MLP & 0.351270 & 0.350988 & 0.349308 & 0.349007 \\
     &  & PINNs & 0.354653 & 0.354871 & 0.350601 & 0.350840 \\
    \cmidrule(lr){2-7}
     & \multirow{6}{*}{0.03} & SDIFT & 0.358035 & 0.380800 & 0.380301 & 0.381514 \\
     &  & LRTFR & 0.311700 & \underline{0.322200} & 0.313404 & \underline{0.320383} \\
     &  & PhysFTM & \textbf{0.017354} & \textbf{0.087100} & \underline{0.050349} & \textbf{0.099881} \\
     &  & RecFNO & \underline{0.018483} & 0.452826 & \textbf{0.021782} & 0.455233 \\
     &  & SIREN-MLP & 0.348683 & 0.349317 & 0.350125 & 0.350761 \\
     &  & PINNs & 0.351663 & 0.352425 & 0.352679 & 0.353309 \\
    \cmidrule(lr){2-7}
     & \multirow{6}{*}{0.15} & SDIFT & 0.294393 & 0.374100 & 0.371273 & 0.371736 \\
     &  & LRTFR & 0.320074 & 0.317200 & 0.319075 & 0.320370 \\
     &  & PhysFTM & \underline{0.029960} & \textbf{0.035900} & \underline{0.057787} & \textbf{0.062454} \\
     &  & RecFNO & \textbf{0.011095} & \underline{0.230709} & \textbf{0.011986} & \underline{0.231022} \\
     &  & SIREN-MLP & 0.350102 & 0.350055 & 0.350040 & 0.349975 \\
     &  & PINNs & 0.353934 & 0.353904 & 0.351765 & 0.351766 \\
    \midrule
    \multirow{18}{*}{Temporal-interval-missing} & \multirow{6}{*}{0.01} & SDIFT & 0.372720 & 0.379670 & 0.372029 & 0.373811 \\
     &  & LRTFR & 0.309972 & \underline{0.330427} & 0.306973 & \underline{0.326716} \\
     &  & PhysFTM & \textbf{0.006558} & \textbf{0.189538} & \underline{0.032476} & \textbf{0.183397} \\
     &  & RecFNO & \underline{0.020743} & 0.701355 & \textbf{0.028592} & 0.681053 \\
     &  & SIREN-MLP & 0.351543 & 0.351296 & 0.342732 & 0.342199 \\
     &  & PINNs & 0.353767 & 0.354026 & 0.347975 & 0.348079 \\
    \cmidrule(lr){2-7}
     & \multirow{6}{*}{0.03} & SDIFT & 0.356991 & 0.361986 & 0.390707 & 0.392586 \\
     &  & LRTFR & 0.310055 & \underline{0.316757} & 0.320590 & \underline{0.327806} \\
     &  & PhysFTM & \textbf{0.016328} & \textbf{0.093969} & \underline{0.039527} & \textbf{0.079239} \\
     &  & RecFNO & \underline{0.018107} & 0.458752 & \textbf{0.021163} & 0.436798 \\
     &  & SIREN-MLP & 0.346053 & 0.346642 & 0.358185 & 0.358978 \\
     &  & PINNs & 0.348742 & 0.349428 & 0.365077 & 0.366217 \\
    \cmidrule(lr){2-7}
     & \multirow{6}{*}{0.15} & SDIFT & 0.294740 & 0.299284 & 0.349127 & 0.350140 \\
     &  & LRTFR & 0.320387 & 0.321854 & 0.305895 & 0.307641 \\
     &  & PhysFTM & \underline{0.031414} & \textbf{0.040888} & \underline{0.045196} & \textbf{0.048007} \\
     &  & RecFNO & \textbf{0.011196} & \underline{0.234129} & \textbf{0.012961} & \underline{0.218094} \\
     &  & SIREN-MLP & 0.353258 & 0.353191 & 0.332458 & 0.332446 \\
     &  & PINNs & 0.356292 & 0.356274 & 0.338695 & 0.338699 \\
    \bottomrule
  \end{tabular}
\end{table}

\begin{table}[p]
\centering
  \scriptsize
  \setlength{\tabcolsep}{3pt}
  \caption{Complete four-subset reconstruction RMSE on NS2D across spatial sensor ratios. Lower is better. (\textbf{Bold}: best, \underline{Underline}: second best)}
  \label{tab:full-reconstruction-ns2d}
  \begin{tabular}{@{}cllrrrr@{}}
    \toprule
    Protocol & $s$ & Method & S-sup & Off-sup & S-miss & Off-miss \\
    \midrule
    \multirow{18}{*}{Random-drop-half} & \multirow{6}{*}{0.01} & SDIFT & 0.518934 & 0.512800 & 0.510791 & 0.509261 \\
     &  & LRTFR & \underline{0.123841} & \underline{0.172000} & \underline{0.126188} & \underline{0.170679} \\
     &  & PhysFTM & \textbf{0.007849} & \textbf{0.137700} & \textbf{0.030927} & \textbf{0.130659} \\
     &  & RecFNO & 0.261520 & 0.418079 & 0.262461 & 0.413223 \\
     &  & SIREN-MLP & 0.244818 & 0.245620 & 0.243905 & 0.244360 \\
     &  & PINNs & 0.338968 & 0.341335 & 0.339094 & 0.341072 \\
    \cmidrule(lr){2-7}
     & \multirow{6}{*}{0.03} & SDIFT & 0.528668 & 0.509700 & 0.510029 & 0.510274 \\
     &  & LRTFR & 0.132450 & 0.147200 & 0.134470 & 0.145757 \\
     &  & PhysFTM & \textbf{0.006687} & \textbf{0.062100} & \textbf{0.032396} & \textbf{0.064976} \\
     &  & RecFNO & \underline{0.068098} & \underline{0.097567} & \underline{0.078196} & \underline{0.104413} \\
     &  & SIREN-MLP & 0.245551 & 0.245255 & 0.244651 & 0.244405 \\
     &  & PINNs & 0.342539 & 0.341771 & 0.341006 & 0.340554 \\
    \cmidrule(lr){2-7}
     & \multirow{6}{*}{0.15} & SDIFT & 0.609226 & 0.510500 & 0.510615 & 0.510589 \\
     &  & LRTFR & 0.135290 & 0.139400 & 0.136582 & 0.138756 \\
     &  & PhysFTM & \textbf{0.005888} & \underline{0.020300} & \textbf{0.032096} & \textbf{0.033357} \\
     &  & RecFNO & \underline{0.007173} & \textbf{0.007929} & \underline{0.039964} & \underline{0.040051} \\
     &  & SIREN-MLP & 0.245407 & 0.245486 & 0.244408 & 0.244476 \\
     &  & PINNs & 0.342011 & 0.342206 & 0.340038 & 0.340193 \\
    \midrule
    \multirow{18}{*}{Temporal-interval-missing} & \multirow{6}{*}{0.01} & SDIFT & 0.522547 & 0.520767 & 0.494061 & 0.492242 \\
     &  & LRTFR & \underline{0.127082} & \underline{0.173888} & \underline{0.179695} & 0.206649 \\
     &  & PhysFTM & \textbf{0.004291} & \textbf{0.141533} & \textbf{0.136380} & \textbf{0.164821} \\
     &  & RecFNO & 0.267930 & 0.428534 & 0.405006 & 0.478397 \\
     &  & SIREN-MLP & 0.249049 & 0.249814 & 0.206410 & \underline{0.206503} \\
     &  & PINNs & 0.346718 & 0.349269 & 0.306146 & 0.306602 \\
    \cmidrule(lr){2-7}
     & \multirow{6}{*}{0.03} & SDIFT & 0.530194 & 0.530518 & 0.504341 & 0.502748 \\
     &  & LRTFR & 0.135148 & 0.146821 & \underline{0.190250} & \underline{0.196613} \\
     &  & PhysFTM & \textbf{0.004346} & \textbf{0.062824} & \textbf{0.149493} & \textbf{0.150558} \\
     &  & RecFNO & \underline{0.069056} & \underline{0.099100} & 0.358033 & 0.359434 \\
     &  & SIREN-MLP & 0.247968 & 0.247773 & 0.213687 & 0.213537 \\
     &  & PINNs & 0.347219 & 0.346569 & 0.318917 & 0.318479 \\
    \cmidrule(lr){2-7}
     & \multirow{6}{*}{0.15} & SDIFT & 0.610478 & 0.609820 & 0.485120 & 0.485171 \\
     &  & LRTFR & 0.138750 & 0.140954 & \underline{0.170839} & \underline{0.172046} \\
     &  & PhysFTM & \textbf{0.005690} & \underline{0.012067} & \textbf{0.134833} & \textbf{0.134567} \\
     &  & RecFNO & \underline{0.007474} & \textbf{0.008253} & 0.318252 & 0.318191 \\
     &  & SIREN-MLP & 0.250208 & 0.250287 & 0.203004 & 0.203190 \\
     &  & PINNs & 0.350376 & 0.350661 & 0.300498 & 0.300163 \\
    \bottomrule
  \end{tabular}
\end{table}

\begin{table}[p]
\centering
  \scriptsize
  \setlength{\tabcolsep}{3pt}
  \caption{Complete four-subset reconstruction RMSE on Heat3D across spatial sensor ratios. Lower is better. (\textbf{Bold}: best, \underline{Underline}: second best)}
  \label{tab:full-reconstruction-heat3d}
  \begin{tabular}{@{}cllrrrr@{}}
    \toprule
    Protocol & $s$ & Method & S-sup & Off-sup & S-miss & Off-miss \\
    \midrule
    \multirow{18}{*}{Random-drop-half} & \multirow{6}{*}{0.01} & SDIFT & 0.058457 & 0.058594 & 0.061825 & 0.062284 \\
     &  & LRTFR & \underline{0.010659} & \underline{0.010815} & \underline{0.012014} & \underline{0.012549} \\
     &  & PhysFTM & \textbf{0.001344} & \textbf{0.001907} & \textbf{0.006379} & \textbf{0.006824} \\
     &  & RecFNO & 0.339625 & 0.888925 & 0.307609 & 0.813064 \\
     &  & SIREN-MLP & 0.040199 & 0.040240 & 0.042215 & 0.042397 \\
     &  & PINNs & 0.040309 & 0.040409 & 0.041993 & 0.042295 \\
    \cmidrule(lr){2-7}
     & \multirow{6}{*}{0.03} & SDIFT & 0.058111 & 0.058081 & 0.063030 & 0.062716 \\
     &  & LRTFR & \underline{0.012121} & \underline{0.012216} & \underline{0.013093} & \underline{0.013146} \\
     &  & PhysFTM & \textbf{0.001083} & \textbf{0.001250} & \textbf{0.005745} & \textbf{0.005744} \\
     &  & RecFNO & 0.287363 & 0.778472 & 0.263722 & 0.720340 \\
     &  & SIREN-MLP & 0.040803 & 0.040865 & 0.042108 & 0.042053 \\
     &  & PINNs & 0.040002 & 0.040113 & 0.042714 & 0.042640 \\
    \cmidrule(lr){2-7}
     & \multirow{6}{*}{0.15} & SDIFT & 0.059938 & 0.059993 & 0.060820 & 0.060836 \\
     &  & LRTFR & \underline{0.012518} & \underline{0.012578} & \underline{0.011607} & \underline{0.011652} \\
     &  & PhysFTM & \textbf{0.001015} & \textbf{0.001040} & \textbf{0.005282} & \textbf{0.005295} \\
     &  & RecFNO & 0.171750 & 0.322953 & 0.158704 & 0.258328 \\
     &  & SIREN-MLP & 0.041383 & 0.041470 & 0.041042 & 0.041105 \\
     &  & PINNs & 0.041778 & 0.041929 & 0.040826 & 0.040917 \\
    \midrule
    \multirow{18}{*}{Temporal-interval-missing} & \multirow{6}{*}{0.01} & SDIFT & 0.059872 & 0.060202 & 0.060550 & 0.060789 \\
     &  & LRTFR & \underline{0.011883} & \underline{0.012311} & \underline{0.012546} & \underline{0.012876} \\
     &  & PhysFTM & \textbf{0.001293} & \textbf{0.002132} & \textbf{0.007193} & \textbf{0.007070} \\
     &  & RecFNO & 0.320509 & 0.837414 & 0.321806 & 0.904206 \\
     &  & SIREN-MLP & 0.040678 & 0.040735 & 0.038204 & 0.038412 \\
     &  & PINNs & 0.040636 & 0.040735 & 0.043391 & 0.044007 \\
    \cmidrule(lr){2-7}
     & \multirow{6}{*}{0.03} & SDIFT & 0.062840 & 0.062680 & 0.050158 & 0.049889 \\
     &  & LRTFR & \underline{0.012100} & \underline{0.012250} & \underline{0.005766} & \underline{0.005767} \\
     &  & PhysFTM & \textbf{0.001218} & \textbf{0.001372} & \textbf{0.003103} & \textbf{0.003046} \\
     &  & RecFNO & 0.286290 & 0.775257 & 0.225969 & 0.641868 \\
     &  & SIREN-MLP & 0.043423 & 0.043440 & 0.027492 & 0.027421 \\
     &  & PINNs & 0.043136 & 0.043190 & 0.033532 & 0.033341 \\
    \cmidrule(lr){2-7}
     & \multirow{6}{*}{0.15} & SDIFT & 0.063097 & 0.063136 & 0.047558 & 0.047574 \\
     &  & LRTFR & \underline{0.011794} & \underline{0.011881} & \underline{0.009763} & \underline{0.009595} \\
     &  & PhysFTM & \textbf{0.001105} & \textbf{0.001136} & \textbf{0.003756} & \textbf{0.003682} \\
     &  & RecFNO & 0.175643 & 0.317359 & 0.122943 & 0.169169 \\
     &  & SIREN-MLP & 0.043686 & 0.043784 & 0.025485 & 0.025478 \\
     &  & PINNs & 0.044045 & 0.044195 & 0.027758 & 0.027711 \\
    \bottomrule
  \end{tabular}
\end{table}

\begin{table}[p]
\centering
  \scriptsize
  \setlength{\tabcolsep}{3pt}
  \caption{Complete four-subset reconstruction RMSE on Ocean across spatial sensor ratios. Lower is better. (\textbf{Bold}: best, \underline{Underline}: second best)}
  \label{tab:full-reconstruction-ocean}
  \begin{tabular}{@{}cllrrrr@{}}
    \toprule
    Protocol & $s$ & Method & S-sup & Off-sup & S-miss & Off-miss \\
    \midrule
    \multirow{18}{*}{Random-drop-half} & \multirow{6}{*}{0.01} & SDIFT & 0.281171 & 0.278458 & 0.280994 & 0.280682 \\
     &  & LRTFR & \underline{0.113447} & \underline{0.126022} & \underline{0.122721} & \underline{0.133850} \\
     &  & PhysFTM & \textbf{0.006599} & \textbf{0.115414} & \textbf{0.076277} & \textbf{0.125325} \\
     &  & RecFNO & 0.758963 & 1.005671 & 0.793739 & 1.002127 \\
     &  & SIREN-MLP & 0.200624 & 0.203220 & 0.196868 & 0.200081 \\
     &  & PINNs & 0.297127 & 0.301674 & 0.295973 & 0.301402 \\
    \cmidrule(lr){2-7}
     & \multirow{6}{*}{0.03} & SDIFT & 0.278627 & 0.280655 & 0.275848 & 0.278376 \\
     &  & LRTFR & \underline{0.110377} & \underline{0.114503} & \underline{0.117666} & \underline{0.121548} \\
     &  & PhysFTM & \textbf{0.006319} & \textbf{0.065738} & \textbf{0.071328} & \textbf{0.088581} \\
     &  & RecFNO & 0.738273 & 0.970741 & 0.776879 & 0.963019 \\
     &  & SIREN-MLP & 0.201582 & 0.201681 & 0.200241 & 0.200943 \\
     &  & PINNs & 0.301625 & 0.300997 & 0.302323 & 0.301937 \\
    \cmidrule(lr){2-7}
     & \multirow{6}{*}{0.15} & SDIFT & 0.275481 & 0.276918 & 0.281502 & 0.282907 \\
     &  & LRTFR & \underline{0.109859} & \underline{0.110456} & \underline{0.127294} & \underline{0.127726} \\
     &  & PhysFTM & \textbf{0.007103} & \textbf{0.017565} & \textbf{0.080431} & \textbf{0.080677} \\
     &  & RecFNO & 0.711010 & 0.854068 & 0.755037 & 0.861364 \\
     &  & SIREN-MLP & 0.199900 & 0.200691 & 0.201670 & 0.202316 \\
     &  & PINNs & 0.298829 & 0.300486 & 0.301210 & 0.302968 \\
    \midrule
    \multirow{18}{*}{Temporal-interval-missing} & \multirow{6}{*}{0.01} & SDIFT & 0.283314 & 0.282061 & 0.272620 & 0.270666 \\
     &  & LRTFR & \underline{0.104156} & \underline{0.119385} & \textbf{0.171227} & \textbf{0.182007} \\
     &  & PhysFTM & \textbf{0.005952} & \textbf{0.115920} & 0.191076 & 0.194097 \\
     &  & RecFNO & 0.763253 & 1.007166 & 0.944159 & 0.992394 \\
     &  & SIREN-MLP & 0.201337 & 0.204394 & \underline{0.182836} & \underline{0.184601} \\
     &  & PINNs & 0.299977 & 0.306116 & 0.282315 & 0.282459 \\
    \cmidrule(lr){2-7}
     & \multirow{6}{*}{0.03} & SDIFT & 0.274940 & 0.277254 & 0.287116 & 0.289410 \\
     &  & LRTFR & \underline{0.105062} & \underline{0.108289} & \textbf{0.180623} & \textbf{0.183770} \\
     &  & PhysFTM & \textbf{0.005946} & \textbf{0.063821} & \underline{0.189849} & \underline{0.185648} \\
     &  & RecFNO & 0.718410 & 0.943793 & 0.991909 & 1.053829 \\
     &  & SIREN-MLP & 0.197733 & 0.198327 & 0.211994 & 0.211505 \\
     &  & PINNs & 0.297650 & 0.297391 & 0.318698 & 0.317248 \\
    \cmidrule(lr){2-7}
     & \multirow{6}{*}{0.15} & SDIFT & 0.277062 & 0.278427 & 0.284215 & 0.285884 \\
     &  & LRTFR & \underline{0.110311} & \underline{0.110758} & 0.190074 & 0.190728 \\
     &  & PhysFTM & \textbf{0.007105} & \textbf{0.017647} & \textbf{0.159473} & \textbf{0.158267} \\
     &  & RecFNO & 0.712010 & 0.855126 & 0.930787 & 0.906187 \\
     &  & SIREN-MLP & 0.202733 & 0.203547 & \underline{0.189757} & \underline{0.190244} \\
     &  & PINNs & 0.300200 & 0.301877 & 0.299327 & 0.301165 \\
    \bottomrule
  \end{tabular}
\end{table}